\documentclass[10pt,twocolumn,letterpaper]{article}
\usepackage[pagenumbers]{cvpr}
\usepackage{multirow}
\usepackage{makecell}
\usepackage[table]{xcolor}
\usepackage{booktabs}
\usepackage{array}
\definecolor{redcolor}{HTML}{C00000}
\definecolor{greencolor}{HTML}{00B050}
\definecolor{cvprblue}{rgb}{0.21,0.49,0.74}
\usepackage[pagebackref,breaklinks,citecolor=cvprblue,urlcolor=cvprblue,linkcolor=cvprblue]{hyperref}
\usepackage{graphicx}

\title{\centering\raisebox{-1.2ex}{\includegraphics[height=2em]{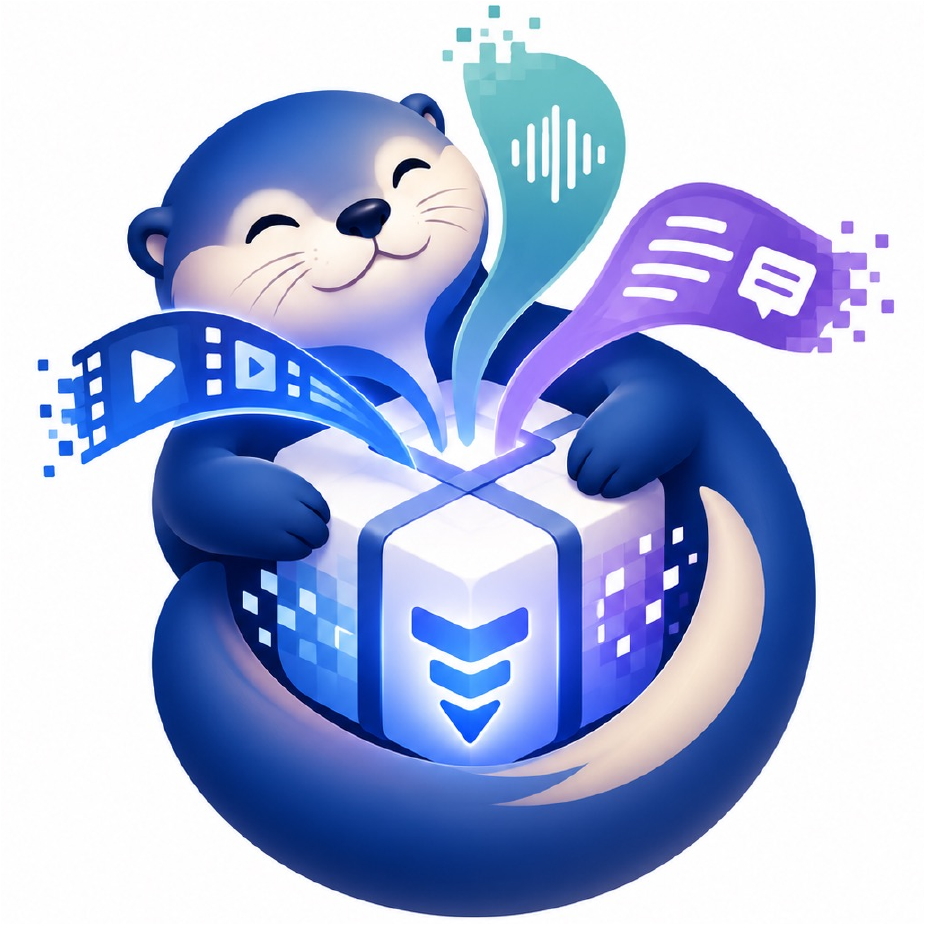}}OmniPack: Unified Token Compression for Efficient\\Omni-modal Large Language Models}
\author{Wanshun Su\textsuperscript{1,*}\quad
    Yang Shi\textsuperscript{2,*}\quad
    Feihu Liu\textsuperscript{3}\quad
    Ziwen Yu\textsuperscript{3}\quad
    Yan Min\textsuperscript{3}\quad
    Zhuoran Zhang\textsuperscript{2}\quad
    Qixun Wang\textsuperscript{2}
    \\
    Haotian Wang\textsuperscript{4}\quad
    Shixuan Liu\textsuperscript{4}\quad
    Yuanxing Zhang\textsuperscript{2}\quad
    Peng Wu\textsuperscript{1,$\ddagger$}\quad
    Chengfu Huo\textsuperscript{3}\quad
    Liang Ding\textsuperscript{3,$\ddagger$}
    \\ 
    \textsuperscript{1}Northwestern Polytechnical University \,
    \textsuperscript{2}Peking University \,
    \textsuperscript{3}Alibaba Group \,
    \textsuperscript{4}Tsinghua University \,
    \\
    \url{https://github.com/RowanSu/OmniPack}
}

\begin{document}
\maketitle

\renewcommand{\thefootnote}{\fnsymbol{footnote}}
\footnotetext[1]{Equal contribution.}
\footnotetext[3]{Corresponding author.}

\begin{abstract}
Omni-modal large language models (Omni-LLMs) have achieved remarkable performance on audio-visual understanding tasks, but processing long and highly redundant visual and audio token sequences incurs substantial computational overhead, demanding aggressive token compression for efficient deployment. 
Existing methods often degrade at low token budgets: pre-LLM compression may discard structurally important and globally distributed evidence, whereas inner-LLM compression often underexploits query-conditioned audio-visual collaboration. 
To address these limitations, we propose \textbf{OmniPack}, a training-free framework that coordinates structural compression before the LLM with task-relevant semantic refinement within the LLM. 
Before the LLM, OmniPack removes structural redundancy through modality-specific importance, global coverage, and similarity-aware merging. 
After sufficient multimodal interaction, it further consolidates diverse, task-relevant representations through textual guidance and audio-visual collaboration. 
Extensive experiments on five benchmarks with three Omni-LLM backbones demonstrate that OmniPack consistently achieves the best performance--efficiency trade-off across diverse retention ratios, outperforming all existing methods. 
Notably, on Qwen2.5-Omni-7B, OmniPack preserves 98.0\% of the original performance while reducing FLOPs to 16.7\%, and still retains 92.9\% of the original performance with only 6.8\% of the original FLOPs. 
\end{abstract} 
\section{Introduction}

The rapid development of omni-modal large language models (Omni-LLMs)~\cite{yang2025humanomniv2,yao2024minicpm,xu2025qwen3}, which jointly model visual, audio, and textual inputs, has significantly advanced unified audio-visual understanding~\cite{fu2024vita,ge2025arc,li2025omnivideobench,shu2025audio,tang2025video,wu2025varcmp,wu2026avadclip}. 
However, densely tokenized visual and audio inputs produce substantially longer sequences than text, incurring considerable computation and memory overhead during inference. 
Efficient deployment therefore requires aggressive token compression that removes redundancy without losing task-relevant multimodal evidence. 

\begin{figure}[t]
  \centering
  \includegraphics[width=1.0\linewidth]{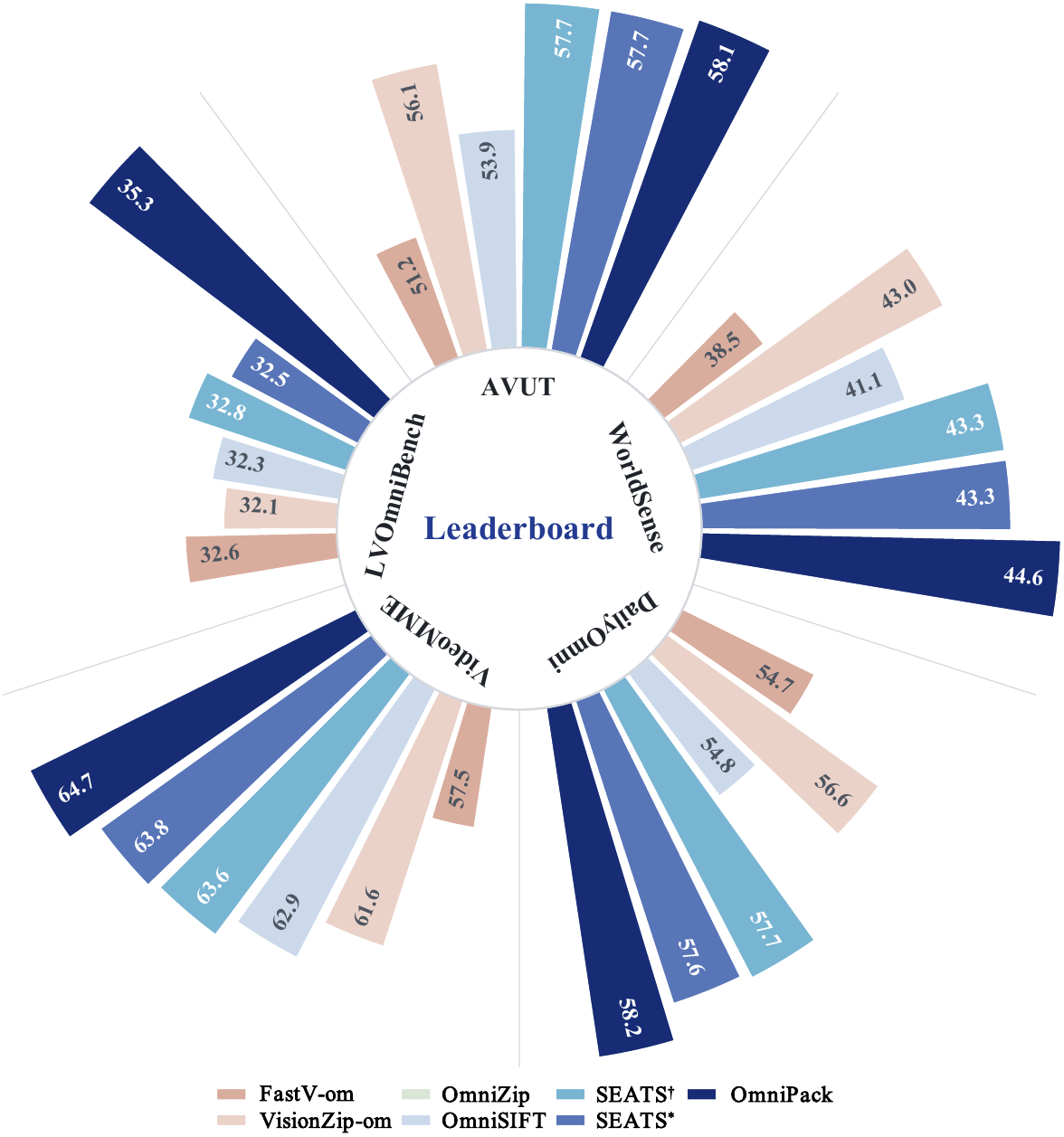}
  \caption{Performance comparison on Qwen2.5-Omni-7B across five benchmarks. SEATS variants and OmniPack further reduce tokens from 15\% to 7.5\%, with OmniPack achieving the best overall performance.}
  \label{fig:visualization_7b_15}
\end{figure}

Audio-visual token compression~\cite{tao2026omnizip,ding2026omnisift} has recently emerged as a practical solution for reducing the prohibitive computational cost of long multimodal sequences. 
Existing Omni-LLM compression methods adopt different compression schedules. 
Some reduce audio-visual tokens only before the LLM~\cite{tao2026omnizip,gong2025echoingpixels,ding2026omnisift,deng2026omnirefine}, whereas others progressively compress tokens both before and within the LLM~\cite{park2026omnidrop,xin2026stage}. 
Despite their effectiveness, existing approaches still face two limitations under aggressive token budgets. 
First, existing methods~\cite{tao2026omnizip,ding2026omnisift} do not sufficiently model the long-range spatiotemporal structure of audio-visual inputs. 
Task-relevant evidence may be sparsely distributed across distant temporal segments, abrupt visual changes, and transient acoustic events. 
Compression based primarily on local importance or token similarity may therefore overlook globally distributed evidence, particularly in long videos. 
Moreover, directly discarding low-scoring tokens can cause irreversible information loss, as individually less salient tokens may still provide complementary context. 
Second, existing methods do not sufficiently align audio-visual collaboration with the semantic maturity of multimodal representations. 
Some methods use one modality to guide the compression of another before the LLM~\cite{tao2026omnizip,ding2026omnisift}. 
At this stage, independently encoded audio and visual representations have undergone limited semantic interaction, such that cross-modal guidance may rely on insufficiently aligned features. 
Although progressive methods further compress tokens within the LLM~\cite{park2026omnidrop,xin2026stage}, their inner-LLM stages primarily rely on textual guidance without explicitly modeling the collaborative interactions between audio and visual tokens. 
These observations motivate a stage-specialized progressive strategy: pre-LLM compression should exploit modality-specific spatiotemporal structure, whereas inner-LLM compression should leverage query-conditioned audio-visual interactions established within the LLM. 

Based on this insight, we propose \textbf{OmniPack}, a training-free two-stage framework that progressively compresses tokens before and within the LLM. 
Pre-LLM compression exploits modality-specific spatiotemporal structure to remove large-scale redundancy while preserving important evidence and global coverage, providing early computational savings. 
After sufficient multimodal interaction, inner-LLM compression further refines the retained representations using textual guidance and audio-visual collaboration. 
Together, the two stages coordinate structural compression with semantic refinement, enabling aggressive token compression while maintaining multimodal understanding performance. 

Extensive experiments demonstrate that OmniPack generalizes effectively across three Omni-LLM backbones and five benchmarks. 
At a final token retention ratio of 7.5\%, OmniPack reduces FLOPs by 10.0$\times$ and achieves a 4.5$\times$ prefill speedup while preserving 95.6\% of the original performance. 
Figure~\ref{fig:visualization_7b_15} further shows that OmniPack consistently offers the best performance--efficiency trade-off among existing token compression methods. 

Our contributions are summarized as follows: 
\begin{itemize}
\item We identify limitations of existing Omni-LLM compression under aggressive token budgets: inadequate long-range audio-visual structure modeling and insufficient exploitation of query-conditioned cross-modal evidence. 
\item We propose OmniPack, a training-free progressive framework that combines modality-specific pre-LLM compression with query-conditioned inner-LLM compression, while merging removed information into retained representatives. 
\item Experiments across multiple Omni-LLMs and benchmarks show that OmniPack preserves strong multimodal understanding while substantially reducing computation and accelerating inference. 
\end{itemize}
\section{Related Work}
\label{sec:formatting}
\subsection{Omni-modal Large Language Models}
Multimodal large language models (MLLMs)~\cite{google2026gemini31pro,openai2026gpt55,shi2025mavors} are rapidly evolving from VLMs~\cite{bai2025qwen3,wang2025internvl3,wang2025monet,wang2026beaconknowingperformagentic,zhang2025debiasing} toward Omni-LLMs~\cite{team2026qwen3,cui2026minicpmo45realtimefullduplex,hurst2024gpt,chen2025avocado}. 
Compared with earlier visual-centric models, Omni-LLMs can seamlessly process and understand text, images, videos, and audio within a unified framework, enabling richer multimodal interactions and broader applications, including audio-visual understanding~\cite{team2026qwen3,chen2025avocado}, real-time conversational assistants~\cite{cui2026minicpmo45realtimefullduplex}, and multimodal agents~\cite{li2026omnigaia}. 
Representative models, such as Gemini 3.1 Pro~\cite{google2026gemini31pro}, GPT-4o~\cite{hurst2024gpt}, Qwen2.5-Omni~\cite{xu2025qwen25omnitechnicalreport}, and MiniCPM-o~\cite{cui2026minicpmo45realtimefullduplex}, have demonstrated strong capabilities in unified multimodal understanding and generation. 
Their ability to jointly model audio and visual cues is particularly valuable for understanding complex, dynamic, and temporally evolving real-world scenarios that require cross-modal reasoning. 
However, dense audio-visual inputs introduce substantial token and computational overhead. 
Therefore, efficient inference is essential for the practical deployment of Omni-LLMs. 

\begin{figure*}[t]
  \centering
  \includegraphics[width=0.95\linewidth]{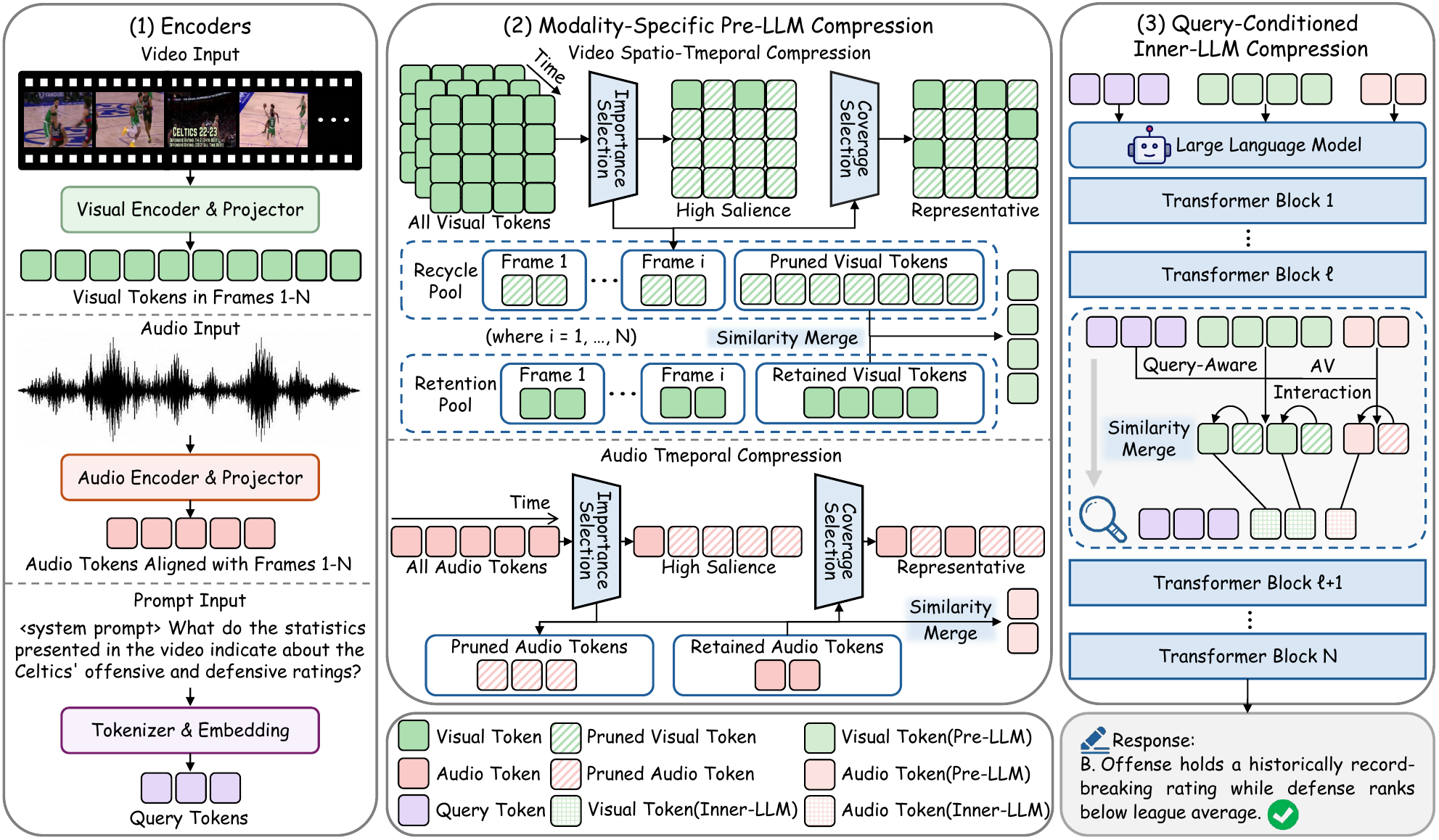}
  \caption{Overview of the OmniPack framework. Before the LLM, modality-specific compression performs importance selection, coverage selection, and similarity-aware token merging. Within the LLM, query-conditioned compression is applied after the $\ell$-th Transformer block. The recycle pool and retention pool denote unselected and selected tokens.}
  \label{fig:framework}
\end{figure*}

\subsection{Token Compression in Omni-LLMs}
Token compression has been extensively studied for image~\cite{bolya2022token,xing2024pyramiddrop,tan2025tokencarve,yang2025visionzip}, video~\cite{tao2025dycoke,shen2026fastvid,fan2026flashvid,du2026unified}, and audio~\cite{lee2025token,lin2025speechprune} inputs to reduce multimodal redundancy and improve inference efficiency. 
With the rapid development of Omni-LLMs, recent studies have extended token compression to joint audio-visual inputs~\cite{tao2026omnizip,ding2026omnisift,gong2026echoingpixelsaliasingresistantjointtoken}. 
OmniZip~\cite{tao2026omnizip} introduces audio-guided video token compression, whereas OmniSIFT~\cite{ding2026omnisift} combines spatio-temporal video compression with vision-guided audio selection. EchoingPixels~\cite{gong2026echoingpixelsaliasingresistantjointtoken}, ContextGuard~\cite{jin2026contextguard}, and OmniRefine~\cite{deng2026omnirefine} investigate joint audio-visual contextualization, cross-modal recoverability, and aligned compression units, respectively. 
OmniDrop~\cite{park2026omnidrop} and SEATS~\cite{xin2026stage} explore progressive, query-guided token pruning within the LLM. 

Despite these advances, existing methods either introduce cross-modal guidance before sufficient audio–visual interaction has been established or perform inner-LLM pruning without explicit audio-visual collaboration. 
As a result, they struggle to jointly preserve modality-specific structure and exploit task-relevant semantics, particularly under aggressive compression. 
This gap motivates a unified framework that coordinates structure-aware pre-LLM compression with query-conditioned audio–visual compression inside the LLM.

\section{Method}
\label{sec:method}

\subsection{Preliminary}
\label{sec:preliminary}

Given a video $\mathcal X^v$ and its aligned audio $\mathcal X^a$, the modality-specific encoder--projector pipelines map them into the LLM embedding space, producing $N_v$ visual tokens $\mathbf X^v=\{\mathbf x_i^v\}_{i=1}^{N_v}$ and $N_a$ audio tokens $\mathbf X^a=\{\mathbf x_i^a\}_{i=1}^{N_a}$. 
Together with the text tokens, they form the multimodal input sequence to the shared LLM backbone. For the encoder tokens 
$\mathbf{X}^{m}=\{\mathbf{x}_{i}^{m}\}_{i=1}^{N_m}$ of modality $m\in\{v,a\}$, we retain $K_m$ tokens before the LLM: 
\begin{equation}
K_m
=
\operatorname{round}
\left(
r_mN_m
\right)
\label{eq:stage1_budget}
\end{equation}
where $N_m$ is the number of input tokens, $r_m$ is the pre-LLM retention ratio, and $K_m$ is the corresponding token budget for modality $m$. 
Let $N_m'=K_m$ denote the number of modality tokens passed to the LLM. 
Given the inner-LLM retention ratio $r_m'$, we further retain the following number of tokens for modality $m$: 
\begin{equation}
K_m'=\operatorname{round}\left(r_m'N_m'\right)
\label{eq:stage2_budget}
\end{equation}
All text tokens remain unchanged throughout compression. 

\subsection{Overview of OmniPack}
\label{sec:overview}

As shown in Figure~\ref{fig:framework}, OmniPack combines modality-specific pre-LLM compression with query-conditioned inner-LLM compression. Pre-LLM compression removes structural redundancy while preserving salient and representative information, whereas inner-LLM compression refines retained tokens using textual relevance, audio-visual collaboration, and within-modality representativeness. 
Together, they shift compression from structural cues to task-conditioned semantics with minimal information loss. 

\subsection{Modality-Specific Pre-LLM Compression}
\label{sec:stage1}

Pre-LLM compression operates independently on each modality, since cross-modal semantics have not yet sufficiently emerged, as illustrated in Figure~\ref{fig:framework}.

\paragraph{Importance Selection.}
We first identify dominant tokens using encoder attention and modality-specific structural variation. 
Attention statistics from the final modality encoder provide a base estimate of token centrality: 
\begin{equation}
\mathbf{A}^{m}
=
\operatorname{Softmax}
\left(
\mathbf{Q}^{m}
(\mathbf{K}^{m})^{\top}
/\sqrt{d}
\right)
\label{eq:encoder_attention}
\end{equation}
Here, $\mathbf{Q}^{m}$ and $\mathbf{K}^{m}$ are the query and key matrices for modality $m$, $d$ is the hidden dimension, and 
$\mathbf{A}^{m}$ is the resulting attention matrix. 
We measure the mean attention received by each token from all other modality tokens, yielding the attention-centrality vector  $\mathbf{a}^{m}\in\mathbb{R}^{N_m}$.

Attention alone may overlook transient structural changes, so we refine it using modality-specific variation cues. 
For video, let $\mathbf{x}_{t,p}^{v}$ denote the token at frame $t$ and spatial position $p$. 
We define adjacent-frame variation and spatial distinctiveness as: 
\begin{equation}
\begin{aligned}
e_{t,p}^{v,\mathrm{temp}}
&=
1-
\cos
\left(
\mathbf{x}_{t,p}^{v},
\mathbf{x}_{t+1,p}^{v}
\right)\\
e_{t,p}^{v,\mathrm{spat}}
&=
1-
\cos
\left(
\mathbf{x}_{t,p}^{v},
\frac{1}{P}
\sum_{q=1}^{P}
\mathbf{x}_{t,q}^{v}
\right)
\end{aligned}
\label{eq:video_structural_cues}
\end{equation}
Here, $P$ is the number of spatial tokens per frame. The two cues capture inter-frame changes and spatial distinctiveness, respectively.

For audio, let $\mathbf{x}_{t}^{a}$ denote the audio token at temporal index $t$.
Adjacent-token variation is defined as: 
\begin{equation}
e_{t}^{a}
=
1-
\cos
\left(
\mathbf{x}_{t}^{a},
\mathbf{x}_{t+1}^{a}
\right)
\label{eq:audio_structural_cue}
\end{equation}
where a larger $e_t^a$ indicates a stronger acoustic boundary or transient event. 

After normalization, the modality-specific structural cues  are added to the attention-centrality vector $\mathbf{a}^{m}$ to obtain the final modality-specific importance vector $\mathbf{s}^{m}=\{s_i^m\}_{i=1}^{N_m}$, where $s_i^m$ denotes the final importance score of token $i$. 
We further denote its normalized score as $\bar{s}_i^m=\mathcal{N}(s_i^m)$. 

Since importance-only selection may overconcentrate tokens around a few salient regions, we allocate only part of the pre-LLM token budget to dominant tokens: 
\begin{equation}
K_m^{\mathrm{imp}}=\operatorname{round}\left(\eta_mK_m\right)
\label{eq:importance_budget}
\end{equation}
where $\eta_m$ denotes the fraction of the token budget allocated to importance selection. 
The top-$K_m^{\mathrm{imp}}$ tokens ranked by $s_i^m$ form the dominant set $\mathcal{S}_{\mathrm{imp}}^m$, while the remaining  $K_m-K_m^{\mathrm{imp}}$ tokens are determined by coverage selection.

\paragraph{Coverage Selection.}
Importance selection may concentrate tokens around a few salient events and leave distant content underrepresented. 
We therefore select additional representatives using both feature and positional distances. 
For modality $m$, we define a joint distance between tokens $\mathbf{x}_i^m$ and $\mathbf{x}_j^m$ as: 
\begin{equation}
d_{ij}^{m} = 1- \cos\left(\mathbf{x}_i^m,\mathbf{x}_j^m\right)+ \lambda d_{ij}^{m,\mathrm{pos}}
\label{eq:stage1_distance}
\end{equation}
where $d_{ij}^{m,\mathrm{pos}}$ denotes the normalized temporal distance for audio or spatiotemporal distance for video. The joint distance captures both representation similarity and structural proximity, preventing distant tokens with similar features from being considered equivalent. 
The parameter $\lambda$ controls the contribution of positional
distance. 

Based on this joint distance, we apply DPC-KNN~\cite{du2016study,rodriguez2014clustering} to identify tokens that are both representative of local neighborhoods and separated from other representative regions. 
The detailed DPC-KNN formulation is provided in the Appendix. 
The top-ranked candidates outside $\mathcal{S}_{\mathrm{imp}}^m$ fill the remaining token budget to form $\mathcal{S}_{\mathrm{cov}}^m$.
The final pre-LLM retained set is: 
\begin{equation}
\mathcal{S}_{\mathrm{pre}}^m = \mathcal{S}_{\mathrm{imp}}^m\cup\mathcal{S}_{\mathrm{cov}}^m,
\quad
\left|\mathcal{S}_{\mathrm{pre}}^m\right|=K_m
\label{eq:stage1_retained_set}
\end{equation}
Thus, importance selection preserves salient evidence, while coverage selection maintains broad spatiotemporal coverage. 

\paragraph{Similarity-Aware Token Merging.}
Instead of directly discarding unselected tokens, we transfer their information to retained representatives. 
Let $\mathcal{U}_{\mathrm{pre}}^m
= \{1,\ldots,N_m\}\setminus\mathcal{S}_{\mathrm{pre}}^m$
denote the set of unselected token indices. 
For each unselected token $i\in\mathcal{U}_{\mathrm{pre}}^m$ and retained token $j\in\mathcal{S}_{\mathrm{pre}}^m$, we compute their affinity and determine the corresponding merge target: 
\begin{equation}
\begin{gathered}
\Phi_{ij}^{m}
=
\cos
\left(
\mathbf{x}_i^m,
\mathbf{x}_j^m
\right)
-
\tau_m d_{ij}^{m,\mathrm{pos}}
+
\zeta\bar{s}_j^m
\\
\pi^m(i)
=
\operatorname*{arg\,max}_{j\in\mathcal{S}_{\mathrm{pre}}^m}
\Phi_{ij}^{m}
\end{gathered}
\label{eq:merge_assignment}
\end{equation}
Here, $\Phi_{ij}^m$ jointly considers feature similarity,  positional proximity, and the normalized importance of retained token $j$. 
The parameter $\tau_m$ controls the modality-specific positional
constraint, while $\zeta$ controls the contribution of representative importance. 
The assignment $\pi^m(i)$ identifies the retained representative that receives unselected token $i$. 

Each retained token is then updated by aggregating the unselected tokens assigned to it: 
\begin{equation}
\widetilde{\mathbf{x}}_j^m = \frac{\mathbf{x}_j^m+\sum_{i:\pi^m(i)=j}
w_i^m\mathbf{x}_i^m}{1+\sum_{i:\pi^m(i)=j}w_i^m},
\quad
j\in\mathcal{S}_{\mathrm{pre}}^m
\label{eq:stage1_merge}
\end{equation}
Here, $w_i^m=(1+\bar{s}_i^m)/2$ assigns larger weights to more important tokens, enabling each representative to aggregate nearby redundant information while mitigating information loss.

The updated representations form $\mathbf{X}_{\mathrm{pre}}^m$, which is fed into the LLM with the text tokens. 

\subsection{Query-Conditioned Inner-LLM Compression}
\label{sec:stage2}

After the first $\ell$ Transformer blocks, the pre-LLM compressed sequence $\mathbf{X}_{\mathrm{pre}}^m$ is transformed into hidden states $\mathbf{H}_{\ell}^{m}=\{\mathbf{h}_{i}^{m}\}_{i=1}^{N_m'}$, where $m\in\{v,a\}$ and $N_m'=K_m$. 
We then perform query-conditioned audio-visual collaborative compression on these representations. 

Let $\mathbf{H}_{\ell}^{q}=\{\mathbf{h}_{j}^{q}\}_{j=1}^{N_q}$ denote the text hidden states. 
We mean-pool $\mathbf{H}_{\ell}^{q}$ into a global query representation $\mathbf{q}$ and the other-modality states $\mathbf{H}_{\ell}^{\bar m}$ into a prototype $\mathbf{p}^{\bar m}$. 
The relevance score of token $\mathbf{h}_i^m$ is: 
\begin{equation}
\begin{aligned}
R_i^m
={}&
\mathcal{N}\left(
\max\left\{
\cos\left(\mathbf{h}_i^m,\mathbf{q}\right),
\max_j \cos\left(\mathbf{h}_i^m,\mathbf{h}_j^q\right)
\right\}
\right)
\\
&+
\mathcal{N}\left(
\cos\left(\mathbf{h}_i^m,\mathbf{p}^{\bar m}\right)
\right)
+
\mathcal{N}\left(u_i^m\right),
\\
u_i^m
={}&
\exp\left(
-\frac{1}{N_m'}
\sum_{j=1}^{N_m'}
\left[
1-\cos\left(\mathbf{h}_i^m,\mathbf{h}_j^m\right)
\right]
\right)
\end{aligned}
\label{eq:inner_llm_score}
\end{equation}
Here, $\mathcal{N}(\cdot)$ denotes modality-wise min--max normalization, and the three terms measure textual relevance, audio-visual collaboration, and within-modality representativeness, respectively. 

We initialize the retained set with the highest-scoring token as: 
$\mathcal{S}_{\mathrm{inner}}^m=\{\operatorname*{arg\,max}_{i}R_i^m\}$.
For each unselected token, we measure its semantic diversity from the retained set by
$D_i^m=\min_{j\in\mathcal{S}_{\mathrm{inner}}^m}[1-\cos(\mathbf{h}_i^m,\mathbf{h}_j^m)]$.
We then select and add the token that best balances relevance and diversity:
\begin{equation}
\begin{gathered}
i^\star
=\operatorname*{arg\,max}_{i\notin\mathcal{S}_{\mathrm{inner}}^m} \left(1+\mathcal{N}\left(R_i^m\right)\right)D_i^m\\
\mathcal{S}_{\mathrm{inner}}^m\leftarrow
\mathcal{S}_{\mathrm{inner}}^m\cup\{i^\star\}
\end{gathered}
\label{eq:inner_llm_selection}
\end{equation}
This process continues until
$|\mathcal{S}_{\mathrm{inner}}^m|=K_m'$.

Each unselected hidden state $\mathbf{h}_i^m$ is assigned to its most cosine-similar retained representative and incorporated through relevance-modulated similarity-weighted averaging. 
The resulting retained states form $\widetilde{\mathbf{H}}_{\ell}^{m}$ with $K_m'$ tokens. 
The compressed audio-visual tokens and all uncompressed text tokens are then passed through the remaining Transformer blocks. 
\section{Experiment}

\subsection{Experimental Settings}

\paragraph{Benchmarks.}
We evaluate OmniPack on five widely used audio-visual understanding
benchmarks: AVUT~\cite{yang2025audio}, WorldSense~\cite{hong2025worldsense}, DailyOmni~\cite{zhou2025dailyomni}, VideoMME~\cite{fu2025videomme}, and LVOmniBench~\cite{tao2026lvomnibench}. 
Together, they cover audio-centric and video-centric understanding, short- and long-form videos, and tasks ranging from perception to reasoning. 

\paragraph{Comparison Methods.}
We compare OmniPack with representative training-free baselines, including image-based FastV~\cite{chen2024image} and VisionZip~\cite{yang2025visionzip}, video-based FastVID~\cite{shen2026fastvid} and VidCom$^2$~\cite{liu2025video}, omni-modal OmniZip~\cite{tao2026omnizip}, OmniSIFT~\cite{ding2026omnisift},  SEATS~\cite{xin2026stage}, and Random. 
Following prior work~\cite{xin2026stage}, FastV-om and VisionZip-om extend FastV and VisionZip to both video and audio tokens. OmniSIFT\textsuperscript{$\circ$} retains only its inference-time compression strategy without alignment training. SEATS\textsuperscript{$\dagger$} uses the original configuration, while SEATS\textsuperscript{$\star$} disables late-layer token removal; both match the final modality-specific retention ratios and are evaluated only on the 28-layer Qwen2.5-Omni-7B and MiniCPM-o-2.6 backbones.

\paragraph{Implementation Details.}
We implement OmniPack on Qwen2.5-Omni-3B/7B~\cite{xu2025qwen25omnitechnicalreport} and MiniCPM-o-2.6~\cite{yao2024minicpm} using NVIDIA H20 GPUs. 
The maximum number of frames is set to 128 for AVUT, WorldSense, and DailyOmni, and 768 for VideoMME and LVOmniBench. 
Following prior work~\cite{tao2026omnizip,xin2026stage}, we evaluate four pre-LLM retention ratios---25\%, 20\%, 15\%, and 10\%---using model-specific visual and audio allocations. 
For inner-LLM compression, we retain 50\% of the pre-LLM tokens at layer 18 for Qwen2.5-Omni-7B and MiniCPM-o-2.6, and layer 26 for Qwen2.5-Omni-3B. 
We set $(\eta_v,\eta_a)$=(0.25,0.35) and $\lambda$=0.20 for coverage selection. For token merging, we use $(\tau_v,\tau_a)$=(0.10,0.05) and $\zeta$=0.10. 
All evaluations are conducted with LMMs-Eval~\cite{zhang2025lmms}. 
Additional details are provided in the Appendix.

\begin{table*}[htbp]
\centering
\setlength{\tabcolsep}{3.2pt}
\renewcommand{\arraystretch}{0.92}
\small
\begin{tabular}{@{}lccc*{5}{c}cc@{}}
\toprule
\multirow{2}{*}{\textbf{Method}}
& \multicolumn{3}{c}{\textbf{Efficiency}}
& \multicolumn{5}{c}{\textbf{Benchmark Performance}}
& \multicolumn{2}{c}{\textbf{Average}} \\
\cmidrule(lr){2-4}
\cmidrule(lr){5-9}
\cmidrule(lr){10-11}
&
\makecell{\textbf{FLOPs (T)}}
&
\makecell{\textbf{Relative}\\\textbf{FLOPs}}
&
\makecell{\textbf{Retention}\\\textbf{Ratio}}
&
\textbf{AVUT}
&
\makecell{\textbf{World}\\\textbf{Sense}}
&
\makecell{\textbf{Daily}\\\textbf{Omni}}
&
\makecell{\textbf{Video}\\\textbf{MME}}
&
\makecell{\textbf{LVOmni}\\\textbf{Video}}
&
\makecell{\textbf{Avg.}\\\textbf{Score}}
&
\makecell{\textbf{Relative}\\\textbf{Perf. (\%)}}
\\
\midrule

\rowcolor{gray!18}
Qwen2.5-Omni-7B
& 73.2 & 100.0\% & 100\% / 100\%
& 64.0 & 46.6 & 63.9 & 64.4 & 34.2
& 54.6 & 100.0 \\

\midrule

Random
& 14.9 & 20.4\% & 25\% / --
& 58.7 & 43.1 & 57.1 & 65.1 & 32.1
& 51.2 & 93.8 \\

FastV
& 19.1 & 26.1\% & 100\% / 25\%
& 54.9 & 42.0 & 56.6 & 60.7 & 33.1
& 49.5 & 90.7 \\

FastV-om
& 19.1 & 26.1\% & 100\% / 25\%
& 59.3 & 43.3 & 58.6 & 64.0 & 32.6
& 51.6 & 94.5 \\

VisionZip
& 14.9 & 20.4\% & 25\% / --
& 56.8 & 44.0 & 59.3 & 63.4 & 31.7
& 51.0 & 93.4 \\

VisionZip-om
& 14.9 & 20.4\% & 25\% / --
& 58.7 & 44.7 & 58.2 & 65.1 & 32.9
& 51.9 & 95.1 \\

FastVID
& 13.6 & 18.5\% & 25\% / --
& 58.0 & 42.8 & 58.2 & 60.1 & 33.2
& 50.5 & 92.5 \\

VidCom$^2$
& 14.9 & 20.4\% & 25\% / --
& 58.8 & 43.3 & 57.7 & 63.7 & 33.1
& 51.3 & 94.0 \\

OmniZip
& 14.9 & 20.4\% & 25\% / --
& 56.9 & 42.6 & 55.9 & 65.9 & \underline{35.0}
& 51.3 & 93.8 \\

OmniSIFT\textsuperscript{$\circ$}
& 14.9 & 20.4\% & 25\% / --
& 58.4 & 42.1 & 58.0 & 65.9 & 32.5
& 51.4 & 94.1 \\

SEATS\textsuperscript{$\dagger$}
& 11.6 & 15.8\% & 25\% / 12.5\%
& 60.0 & 45.0 & 59.7 & \underline{66.3} & 33.5
& 52.9 & 96.7 \\

SEATS\textsuperscript{$\star$}
& 12.5 & 17.1\% & 25\% / 12.5\%
& 60.1 & 45.0 & 59.8 & \textbf{66.4} & 33.9
& 53.0 & 97.1 \\

\addlinespace[1pt]
\textbf{OmniPack (w/o M)}
& 14.9 & 20.4\% & 25\% / --
& \underline{60.7} & \textbf{45.3} & \underline{60.3}
& 66.0 & \textbf{35.6}
& \textbf{53.6} & \textbf{98.2} \\

\rowcolor{blue!8}
\textbf{OmniPack}
& 12.2 & 16.7\% & 25\% / 12.5\%
& \textbf{61.0} & \textbf{45.3} & \textbf{60.6}
& 65.7 & \underline{35.0}
& \underline{53.5} & \underline{98.0} \\

\midrule

Random
& 11.8 & 16.2\% & 20\% / --
& 58.6 & 42.7 & 56.7 & 64.0 & \underline{34.3}
& 51.3 & 94.0 \\

FastV-om
& 16.2 & 22.2\% & 100\% / 20\%
& 56.2 & 41.6 & 56.1 & 62.3 & 31.9
& 49.6 & 90.8 \\

VisionZip
& 11.8 & 16.2\% & 20\% / --
& 56.8 & 43.4 & 57.7 & 61.6 & 31.7
& 50.2 & 91.9 \\

VisionZip-om
& 11.8 & 16.2\% & 20\% / --
& 57.1 & 44.0 & 58.0 & 64.3 & 32.6
& 51.2 & 93.8 \\

FastVID
& 11.2 & 15.3\% & 20\% / --
& 55.9 & 42.2 & 56.4 & 59.1 & 32.5
& 49.2 & 90.1 \\

VidCom$^2$
& 11.8 & 16.2\% & 20\% / --
& 55.6 & 41.7 & 55.1 & 59.9 & 32.4
& 48.9 & 89.6 \\

OmniSIFT\textsuperscript{$\circ$}
& 11.8 & 16.2\% & 20\% / --
& 56.5 & 42.2 & 54.3 & 64.7 & 33.3
& 50.2 & 91.9 \\

SEATS\textsuperscript{$\dagger$}
& 9.2 & 12.6\% & 20\% / 10\%
& 59.1 & 44.2 & 58.6 & \underline{65.3} & 33.9
& 52.2 & 95.6 \\

SEATS\textsuperscript{$\star$}
& 10.0 & 13.6\% & 20\% / 10\%
& 59.1 & 44.5 & \textbf{58.9} & \textbf{65.4} & 33.8
& \underline{52.3} & \underline{95.7} \\

\addlinespace[1pt]
\textbf{OmniPack (w/o M)}
& 11.8 & 16.2\% & 20\% / --
& \textbf{60.0} & \textbf{45.0} & \underline{58.7}
& 65.0 & \textbf{35.7}
& \textbf{52.9} & \textbf{96.9} \\

\rowcolor{blue!8}
\textbf{OmniPack}
& 9.8 & 13.3\% & 20\% / 10\%
& \textbf{60.0} & \underline{44.8} & 58.6
& \underline{65.3} & \textbf{35.7}
& \textbf{52.9} & \textbf{96.9} \\

\midrule

Random
& 8.9 & 12.1\% & 15\% / --
& 56.4 & 41.1 & 54.8 & 63.0 & 33.3
& 49.7 & 91.0 \\

FastV-om
& 13.5 & 18.4\% & 100\% / 15\%
& 51.2 & 38.5 & 54.7 & 57.5 & 32.6
& 46.9 & 85.9 \\

VisionZip-om
& 8.9 & 12.1\% & 15\% / --
& 56.1 & 43.0 & 56.6 & 61.6 & 32.1
& 49.9 & 91.4 \\

OmniSIFT\textsuperscript{$\circ$}
& 8.9 & 12.1\% & 15\% / --
& 53.9 & 41.1 & 54.8 & 62.9 & 32.3
& 49.0 & 89.7 \\

SEATS\textsuperscript{$\dagger$}
& 6.9 & 9.5\% & 15\% / 7.5\%
& 57.7 & 43.3 & \underline{57.7} & 63.6 & 32.8
& 51.0 & 93.4 \\

SEATS\textsuperscript{$\star$}
& 7.5 & 10.3\% & 15\% / 7.5\%
& 57.7 & 43.4 & 57.6 & \underline{63.8} & 32.5
& 51.0 & 93.4 \\

\addlinespace[1pt]
\textbf{OmniPack (w/o M)}
& 8.9 & 12.1\% & 15\% / --
& \underline{57.8} & \textbf{44.6} & \textbf{58.2}
& \textbf{64.7} & \underline{34.9}
& \underline{52.0} & \underline{95.2} \\

\rowcolor{blue!8}
\textbf{OmniPack}
& 7.3 & 10.0\% & 15\% / 7.5\%
& \textbf{58.1} & \textbf{44.6} & \textbf{58.2}
& \textbf{64.7} & \textbf{35.3}
& \textbf{52.2} & \textbf{95.6} \\

\midrule

Random
& 6.0 & 8.2\% & 10\% / --
& 53.4 & 39.3 & 52.0 & 61.1 & \underline{33.7}
& 47.9 & 87.7 \\

VisionZip-om
& 6.0 & 8.2\% & 10\% / --
& 53.6 & 41.4 & \underline{55.0} & 60.0 & 33.4
& 48.7 & 89.2 \\

OmniSIFT\textsuperscript{$\circ$}
& 6.0 & 8.2\% & 10\% / --
& 51.8 & 38.8 & 51.4 & 61.3 & 32.8
& 47.2 & 86.4 \\

SEATS\textsuperscript{$\dagger$}
& 4.7 & 6.5\% & 10\% / 5\%
& \underline{56.7} & 42.0 & 55.3 & 62.0 & 32.9
& 49.8 & 91.2 \\

SEATS\textsuperscript{$\star$}
& 5.1 & 7.0\% & 10\% / 5\%
& \underline{56.7} & 42.3 & 55.5 & 61.7 & 32.8
& 49.8 & 91.2 \\

\addlinespace[1pt]
\textbf{OmniPack (w/o M)}
& 6.0 & 8.2\% & 10\% / --
& \textbf{56.8} & \textbf{42.9} & \textbf{56.1}
& \underline{63.5} & \textbf{34.8}
& \textbf{50.8} & \textbf{93.0} \\

\rowcolor{blue!8}
\textbf{OmniPack}
& 5.0 & 6.8\% & 10\% / 5\%
& 56.4 & \underline{42.7} & \textbf{56.1}
& \textbf{63.6} & \textbf{34.8}
& \underline{50.7} & \underline{92.9} \\

\bottomrule
\end{tabular}

\caption{Comparison with state-of-the-art methods on Qwen2.5-Omni-7B. The A\%/B\% denotes the retention ratios before and after inner-LLM compression, respectively. ``--'' indicates that inner-LLM compression is not applied. Best results are in bold, and second-best results are underlined. ``OmniPack (w/o M)'' denotes our method without inner-LLM compression.}
\label{tab:main_results_7b}
\end{table*}

\subsection{Main Results}

\paragraph{State-of-the-Art Performance.}
As shown in Table~\ref{tab:main_results_7b}, our approach  achieves state-of-the-art performance across all retention settings. 
At a 25\% retention ratio, OmniPack (w/o M) preserves 98.2\% of the original performance. 
When the retention ratio is reduced to 15\%, VisionZip-om and 
OmniSIFT\textsuperscript{$\circ$} preserve 91.4\% and 89.7\% of the original performance, respectively, whereas OmniPack (w/o M) retains 95.2\%. 
Under extreme compression at a 10\% retention ratio, OmniPack (w/o M) still preserves 93.0\% of the original performance using only  8.2\% of the original FLOPs, outperforming both SEATS variants and VisionZip-om. 
The query-conditioned inner-LLM compression further reduces computation while maintaining comparable performance. 
At the 15\%/7.5\% and 10\%/5\% retention settings, OmniPack preserves 95.6\% and 92.9\% of the original performance while using only 10.0\% and 6.8\% of the original FLOPs, respectively.
These results demonstrate a strong performance--efficiency trade-off under extremely low token budgets. 

\paragraph{Different Backbones.}
\begin{table*}[htbp]
\centering
\setlength{\tabcolsep}{3.2pt}
\renewcommand{\arraystretch}{0.92}
\small
\begin{tabular}{@{}lccc*{5}{c}cc@{}}
\toprule
\multirow{2}{*}{\textbf{Method}}
& \multicolumn{3}{c}{\textbf{Efficiency}}
& \multicolumn{5}{c}{\textbf{Benchmark Performance}}
& \multicolumn{2}{c}{\textbf{Average}} \\
\cmidrule(lr){2-4}
\cmidrule(lr){5-9}
\cmidrule(lr){10-11}
&
\makecell{\textbf{FLOPs (T)}}
&
\makecell{\textbf{Relative}\\\textbf{FLOPs}}
&
\makecell{\textbf{Retention}\\\textbf{Ratio}}
&
\textbf{AVUT}
&
\makecell{\textbf{World}\\\textbf{Sense}}
&
\makecell{\textbf{Daily}\\\textbf{Omni}}
&
\makecell{\textbf{Video}\\\textbf{MME}}
&
\makecell{\textbf{LVOmni}\\\textbf{Bench}}
&
\makecell{\textbf{Avg.}\\\textbf{Score}}
&
\makecell{\textbf{Relative}\\\textbf{Perf. (\%)}}
\\
\midrule

\rowcolor{gray!18}
Qwen2.5-Omni-3B
& 37.4 & 100.0\% & 100\% / 100\%
& 62.5 & 46.5 & 61.0 & 62.5 & 35.2
& 53.5 & 100.0 \\

\midrule

Random
& 3.9 & 10.5\% & 15\% / --
& 54.4 & 40.4 & 50.5 & 59.4 & 32.6
& 47.5 & 88.8 \\

FastV-om
& 5.8 & 15.5\% & 100\% / 15\%
& 51.0 & 41.1 & 50.5 & 53.4 & 32.6
& 45.7 & 85.4 \\

VisionZip-om
& 3.9 & 10.5\% & 15\% / --
& 54.2 & 42.9 & \textbf{54.1} & 59.6 & 32.3
& \underline{48.6} & \underline{90.8} \\

OmniSIFT\textsuperscript{$\circ$}
& 3.9 & 10.5\% & 15\% / --
& 52.9 & 39.4 & 49.4 & 59.5 & \underline{33.9}
& 47.0 & 87.9 \\

\textbf{OmniPack (w/o M)}
& 3.9 & 10.5\% & 15\% / --
& \underline{57.1} & \underline{43.8} & \underline{53.2}
& \underline{59.8} & \textbf{34.0}
& \textbf{49.6} & \textbf{92.7} \\

\rowcolor{blue!8}
\textbf{OmniPack}
& 3.4 & 9.0\% & 15\% / 7.5\%
& \textbf{57.4} & \textbf{43.9} & 52.7
& \textbf{60.2} & 33.8
& \textbf{49.6} & \textbf{92.7} \\

\midrule

\rowcolor{gray!18}
MiniCPM-o-2.6
& 73.2 & 100.0\% & 100\% / 100\%
& 53.2 & 40.8 & 52.2 & 62.0 & 29.5
& 47.5 & 100.0 \\

\midrule

Random
& 8.9 & 12.1\% & 15\% / --
& 47.7 & 38.5 & \underline{51.0} & \underline{61.4} & \textbf{36.0}
& 46.9 & 98.7 \\

FastV-om
& 13.5 & 18.4\% & 100\% / 15\%
& 43.5 & 36.1 & 43.8 & 51.6 & 32.7
& 41.5 & 87.4 \\

VisionZip-om
& 8.9 & 12.1\% & 15\% / --
& 49.4 & 38.7 & \textbf{52.0} & 60.2 & \underline{35.5}
& 47.2 & 99.4 \\

OmniSIFT\textsuperscript{$\circ$}
& 8.9 & 12.1\% & 15\% / --
& 48.4 & 37.9 & 50.6 & 59.6 & 34.5
& 46.2 & 97.3 \\

SEATS\textsuperscript{$\dagger$}
& 6.9 & 9.5\% & 15\% / 7.5\%
& 49.8 & 38.8 & 50.5 & 60.0 & 34.4
& 46.7 & 98.3 \\

SEATS\textsuperscript{$\star$}
& 7.5 & 10.3\% & 15\% / 7.5\%
& 49.8 & 39.1 & 49.1 & 60.4 & 34.6
& 46.6 & 98.1 \\

\textbf{OmniPack (w/o M)}
& 8.9 & 12.1\% & 15\% / --
& \textbf{51.3} & \textbf{40.0} & \textbf{52.0}
& 61.1 & 34.3
& \underline{47.7} & \underline{100.4} \\

\rowcolor{blue!8}
\textbf{OmniPack}
& 7.3 & 10.0\% & 15\% / 7.5\%
& \underline{51.2} & \underline{39.8} & \textbf{52.0}
& \textbf{61.7} & 34.6
& \textbf{47.9} & \textbf{100.8} \\

\bottomrule
\end{tabular}
\caption{Cross-backbone comparison on Qwen2.5-Omni-3B and MiniCPM-o-2.6.}
\label{tab:main_results_cross_backbone_15}
\end{table*}
Table~\ref{tab:main_results_cross_backbone_15} further evaluates our method on Qwen2.5-Omni-3B and MiniCPM-o-2.6, covering different model scales and architectures. 
On Qwen2.5-Omni-3B, OmniPack preserves 92.7\% of the original performance at a 15\%/7.5\% retention ratio while using only 9.0\% of the original FLOPs. 
On MiniCPM-o-2.6, OmniPack achieves 100.8\% of the original performance under the same retention setting, outperforming existing compression methods while using only 10.0\% of the original FLOPs. 

\subsection{Efficiency Analysis}
\begin{table}[htbp]
\centering
\setlength{\tabcolsep}{2.5pt}
\resizebox{1.0\linewidth}{!}{
\begin{tabular}{l|c|r|c|c}
\toprule
\textbf{Method} &
\makecell{\textbf{Token Retention}\\\textbf{Pre-LLM / Final}} &
\makecell{\bf FLOPs (T)$\downarrow$} &
\makecell{\bf Prefill \\ \bf (sec.)$\downarrow$} &
\makecell{\bf Avg. \\ \bf Score$\uparrow$} \\
\midrule
\rowcolor{gray!18}
Vanilla
& 100\% & 73.2~$(1.0\times)$ & 1.512~$(1.0\times)$ & 54.6 \\
\midrule
FastV-om
& 100\%/25\% & 19.1\,(3.8$\times$) & 0.415\,(3.6$\times$) & 51.6 \\
OmniZip
& 25\%/-- & 14.9\,(4.9$\times$) & 0.360\,(4.2$\times$) & 51.3 \\
VisionZip-om
& 25\%/-- & 14.9\,(4.9$\times$) & 0.382\,(4.0$\times$) & 51.9 \\
OmniSIFT\textsuperscript{$\circ$}
& 25\%/-- & 14.9\,(4.9$\times$) & 0.434\,(3.5$\times$) & 51.4 \\
SEATS\textsuperscript{$\dagger$}
& 25\%/12.5\% & 11.6\,(6.3$\times$) & 0.348\,(4.3$\times$) & 52.9 \\
SEATS\textsuperscript{$\star$}
& 25\%/12.5\% & 12.5\,(5.9$\times$) & 0.375\,(4.0$\times$) & 53.0 \\
\textbf{OmniPack (w/o M)}
& 25\%/-- & 14.9\,(4.9$\times$) & 0.396\,(3.8$\times$) & \textbf{53.6} \\
\textbf{OmniPack}
& 25\%/12.5\% & 12.2\,(6.0$\times$) & 0.504\,(3.0$\times$) & 53.5 \\
\textbf{OmniPack}
& 20\%/10\% & 9.8\,(7.5$\times$) & 0.424\,(3.6$\times$) & 52.9 \\
\rowcolor{blue!8}
\textbf{OmniPack}
& 15\%/7.5\% & 7.3\,(\textbf{10.0$\times$}) & 0.339\,(\textbf{4.5$\times$}) & 52.2 \\
\bottomrule
\end{tabular}}
\caption{Efficiency analysis on Qwen2.5-Omni-7B using an NVIDIA H20 GPU.}
\label{table:efficiency}
\end{table}

As shown in Table~\ref{table:efficiency}, at the 25\% retention ratio, OmniPack (w/o M) preserves 98.2\% of the original performance while reducing FLOPs by 4.9$\times$. Crucially, at the 15\%/7.5\% retention ratio, OmniPack preserves 95.6\% of the original performance, outperforming all pre-LLM compression methods at 25\% retention ratio despite retaining substantially fewer tokens. 
This setting reduces FLOPs by 10.0$\times$ and achieves a 4.5$\times$ prefill speedup, demonstrating a favorable performance--efficiency trade-off. 

\subsection{Ablation Studies}
We conduct ablation studies on our pre-LLM compression strategy, our inner-LLM compression strategy, and their compatibility. 
All ablation experiments are conducted on Qwen2.5-Omni-7B. 
Pre-LLM compression strategy ablations are evaluated at a 15\% retention ratio without inner-LLM compression. Unless otherwise specified, the remaining ablations use a 15\%/7.5\% retention setting. 

\paragraph{Pre-LLM Compression Strategy.}
\begin{table}[htbp]
\centering
\resizebox{\linewidth}{!}{
\begin{tabular}{@{}ccc|ccc@{}}
\toprule
\makecell{\bf Importance} &
\makecell{\bf Coverage} &
\makecell{\bf Merge} &
\makecell{\bf World\\\bf Sense} &
\makecell{\bf Video\\\bf MME} &
\makecell{\bf LVOmni\\\bf Bench} \\
\midrule
\checkmark &            &            & 41.7 & 62.0 & 32.7 \\
           & \checkmark &            & 43.1 & 63.3 & 34.7 \\
           &            & \checkmark & 41.7 & 63.9 & 33.7 \\
\checkmark & \checkmark &            & 44.4 & 64.5 & 33.3 \\
\checkmark &            & \checkmark & 41.9 & 62.4 & 32.1 \\
           & \checkmark & \checkmark & 44.2 & 62.7 & 34.1 \\
\checkmark & \checkmark & \checkmark
           & \textbf{44.6} & \textbf{64.7} & \textbf{34.9} \\
\bottomrule
\end{tabular}}
\caption{Ablation of the pre-LLM compression components.}
\label{tab:stage1_module_ablation}
\end{table}

\begin{table}[htbp]
\centering
\resizebox{0.8\linewidth}{!}{
\begin{tabular}{@{}l|ccc@{}}
\toprule
\textbf{Method} &
\textbf{AVUT} &
\makecell{\textbf{World}\\\textbf{Sense}} &
\makecell{\textbf{Daily}\\\textbf{Omni}} \\
\midrule
w/o A-V Collaboration & 57.8 & 44.5 & 57.8 \\
w/ A-V Collaboration & \textbf{58.1} & \textbf{44.6} & \textbf{58.2} \\
\bottomrule
\end{tabular}}
\caption{Ablation of audio-visual collaboration.}
\label{tab:av_collaboration}
\end{table}
\begin{table}[htbp]
\centering
\resizebox{\linewidth}{!}{
\begin{tabular}{@{}l|ccc@{}}
\toprule
\textbf{Guidance Strategy} &
\textbf{AVUT} &
\makecell{\textbf{World}\\\textbf{Sense}} &
\makecell{\textbf{Daily}\\\textbf{Omni}} \\
\midrule
General Query & 57.7 & \textbf{44.6} & 58.1 \\
Last-Token Attention & 57.9 & 44.5 & \textbf{58.2} \\
OmniPack (Text-Aware Guidance) & \textbf{58.1} & \textbf{44.6} & \textbf{58.2} \\
\bottomrule
\end{tabular}}
\caption{Ablation of different text-guidance mechanisms.}
\label{tab:text_guidance}
\end{table}
\begin{table*}[htbp]
\centering
\small
\setlength{\tabcolsep}{3pt}
\renewcommand{\arraystretch}{0.88}
\setlength{\aboverulesep}{0.35ex}
\setlength{\belowrulesep}{0.35ex}
\setlength{\abovetopsep}{0pt}
\setlength{\belowbottomsep}{0pt}
\begin{tabular}{@{}c|cc|ccccc|c@{}}
\toprule
\textbf{ID}
&
\makecell{\textbf{Pre-LLM}}
&
\makecell{\textbf{Inner-LLM}}
&
\makecell{\textbf{AVUT}}
&
\makecell{\textbf{World}\\\textbf{Sense}}
&
\makecell{\textbf{Daily}\\\textbf{Omni}}
&
\makecell{\textbf{Video}\\\textbf{MME}}
&
\makecell{\textbf{LVOmni}\\\textbf{Bench}}
&
\makecell{\textbf{Avg.}\\\textbf{Score}}
\\
\midrule

Vanilla & -- & -- & 64.0 & 46.6 & 63.9 & 64.4 & 34.2 & 54.6 \\
\midrule
1 & VisionZip-om & Ours & 56.2 & 42.9 & 56.6 & 61.7 & 31.1 & 49.7 \\
2 & OmniSIFT\textsuperscript{$\circ$} & Ours & 53.5 & 41.0 & 53.7
& 64.0 & 32.3 & 48.9 \\
3 & SEATS\textsuperscript{$\dagger$} & Ours & 57.4 & 43.1 & 57.4 & 63.7 & 33.1 & 50.9 \\
4 & Ours & SEATS\textsuperscript{$\star$} & 58.0 & \textbf{44.7} & 57.8 & \textbf{64.7} & 33.9 & 51.8 \\
OmniPack & Ours & Ours & \textbf{58.1} & 44.6 & \textbf{58.2} & \textbf{64.7} & \textbf{35.3} & \textbf{52.2} \\

\bottomrule
\end{tabular}
\caption{Compatibility of pre-LLM and inner-LLM compression strategies.}
\label{tab:stage_wise_selection_ablation}
\end{table*}
As shown in Table~\ref{tab:stage1_module_ablation}, coverage selection performs best among the individual components on WorldSense and LVOmniBench. Relative to importance selection, it improves the two benchmarks by approximately 3.4\% and 6.1\%, respectively. Relative to token merging, the gains are 3.4\% and 3.0\%. Combining importance and coverage further improves WorldSense and VideoMME over coverage alone by approximately 3.0\% and 1.9\%. Incorporating all three components achieves the best performance across the three benchmarks, including a further 4.8\% gain on LVOmniBench over the importance-and-coverage configuration. These results demonstrate the complementary roles of the three components in pre-LLM compression.

\paragraph{Inner-LLM Compression Strategy.}

\begin{figure}[t]
  \centering
  \includegraphics[width=0.8\linewidth]{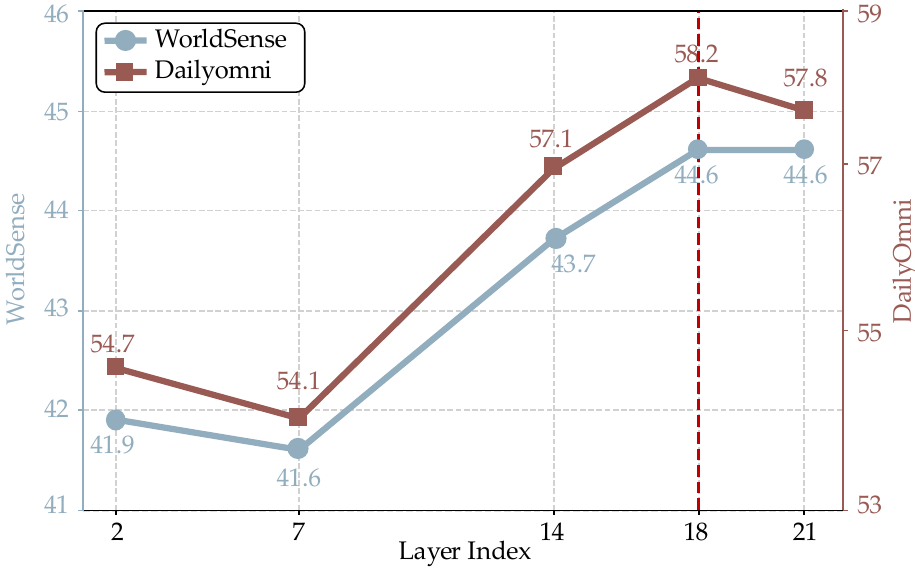}
  \caption{Effect of the inner-LLM compression layer.}
  \label{fig:selection_layer}
\end{figure}
Figure~\ref{fig:selection_layer} and Tables~\ref{tab:av_collaboration} and~\ref{tab:text_guidance} analyze the key mechanisms of our inner-LLM compression strategy. 
Applying the strategy too early limits task-conditioned multimodal interaction, whereas applying it too late reduces the computational benefit. Layer 18 achieves the best overall performance and is therefore adopted by default. 
Audio-visual collaboration consistently outperforms independent compression by better preserving complementary cross-modal information. 
Moreover, compared with question-agnostic general query and last-token guidance, the proposed text-aware guidance aggregates richer semantic cues from the full textual query and achieves the best overall performance. 

\paragraph{Compression Strategies Compatibility.}
Table~\ref{tab:stage_wise_selection_ablation} evaluates the compatibility of different pre-LLM and inner-LLM compression strategies. 
Pairing our inner-LLM compression with VisionZip-om, OmniSIFT\textsuperscript{$\circ$}, or SEATS\textsuperscript{$\dagger$} consistently underperforms the full OmniPack, while replacing our inner-LLM compression with SEATS\textsuperscript{$\star$} also lowers the average score and degrades three of the five benchmarks. 
These results demonstrate that the two stage-specific strategies are complementary and work effectively together. 
\section{Conclusion}
We present OmniPack, a training-free token compression framework for efficient Omni-LLMs. 
Rather than restricting compression to a single location, OmniPack progressively exploits modality-specific structural information before the LLM and task-conditioned semantic information after sufficient multimodal interaction within the LLM. 
This progressive design enables substantially more aggressive compression while effectively preserving downstream performance.
Extensive experiments on five audio-visual understanding benchmarks and three Omni-LLM backbones demonstrate that OmniPack consistently achieves the best performance--efficiency trade-off among existing methods across diverse token budgets. 

{
    \small
    \bibliographystyle{ieeenat_fullname}
    \bibliography{main}
}

\appendix
\clearpage
\setcounter{page}{1}
\maketitlesupplementary

\section{Benchmark Details}
\label{app:benchmark_details}
\begin{table}[htbp]
\centering
\resizebox{1.0\linewidth}{!}{
\begin{tabular}{@{}l rr r@{}}
  \toprule
  \textbf{Benchmark} & \textbf{\#Videos} & \textbf{\#QA Pairs} & \textbf{Duration(sec)} \\
  \midrule
    AVUT~\cite{yang2025audio}\textsuperscript{\#}  & 691 & 1,734 & 69.1 \\
    WorldSense~\cite{hong2025worldsense} & 1,662  & 3,172 & 140.7 \\
    DailyOmni~\cite{zhou2025dailyomni} & 684 & 1,197 & 43.2 \\
    VideoMME~\cite{fu2025videomme} & 900 & 2,700 & 1017.9 \\
    LVOmniBench~\cite{tao2026lvomnibench} & 275 & 1,014 & 2069.7 \\
  \bottomrule
\end{tabular}}
\begin{minipage}{1.0\linewidth} 
\footnotesize \textsuperscript{\#} The full AVUT benchmark contains 2,662 videos and 11,609 QA pairs. Following the evaluation setting of OmniZip~\cite{tao2026omnizip}, we use only the human-annotated subset, which contains 691 videos and 1,734 QA pairs. 
\end{minipage}
\caption{Statistics of the five benchmarks used in our experiments. For each benchmark, we report the number of videos, QA pairs, and the average video duration.}
\label{table:benchmark}
\end{table}
Table~\ref{table:benchmark} summarizes the five benchmarks used in our evaluation. 
Together, they contain 4,212 videos and 9,817 QA pairs, covering diverse real-world scenarios and a wide range of temporal scales. 
These benchmarks differ substantially in their dependence on audio and visual evidence, enabling a comprehensive assessment of token compression across different durations, modalities, and reasoning requirements. 

In what follows, we briefly introduce each benchmark.

\noindent\textbf{AVUT}~\cite{yang2025audio} is an audio-centric video understanding benchmark designed to evaluate both audio-content understanding and audio-visual alignment while minimizing textual shortcuts. 
Its questions cover tasks such as audio information extraction, audio-event localization, and matching auditory events with their corresponding visual entities or actions. 
Following OmniZip~\cite{tao2026omnizip}, we evaluate on its human-annotated subset, which contains 691 videos and 1,734 QA pairs, with an average duration of 69.1 seconds. 

\noindent\textbf{WorldSense}~\cite{hong2025worldsense} comprises 1,662 synchronized audio-visual videos spanning 8 major domains and 67 fine-grained categories, together with 3,172 expert-annotated multiple-choice questions across 26 tasks. 
Its questions are designed with strong audio-visual coupling, requiring models to jointly integrate evidence from both modalities rather than relying on either modality alone. 
The videos have an average duration of 140.7 seconds.

\noindent\textbf{DailyOmni}~\cite{zhou2025dailyomni} consists of 684 videos collected from diverse daily-life scenarios and 1,197 multiple-choice QA pairs across 6 major tasks. 
The benchmark emphasizes temporally aligned audio-visual reasoning, ranging from basic event alignment and temporal ordering to cross-modal inference and contextual reasoning. 
Its videos contain rich auditory and visual events and have an average duration of 43.2 seconds. 

\noindent\textbf{VideoMME}~\cite{fu2025videomme} contains 900 videos covering 6 major domains, with durations ranging from 11 seconds to 1 hour. 
The videos are divided into short, medium, and long subsets and are accompanied by 2,700 human-annotated multiple-choice QA pairs. 
VideoMME supports audio inputs, allowing us to evaluate token compression under different modality configurations. 
Its average video duration is 1,017.9 seconds.

\noindent\textbf{LVOmniBench}~\cite{tao2026lvomnibench} is a long-form audio-visual understanding benchmark containing 275 videos and 1,014 multiple-choice questions. 
Its videos range from 10 to 90 minutes and have an average duration of 2,069.7 seconds. 
The benchmark places particular emphasis on long-range audio-visual correspondence, multi-event information retention, and temporal localization, making it well suited for evaluating compression methods under extremely long multimodal contexts.

\section{More Implementation Details}
\label{app:more_experimental_details}
\begin{table}[htbp]
\centering
\resizebox{1.0\linewidth}{!}{
\begin{tabular}{c cc cc}
\toprule
\multirow[c]{2}{*}[-0.6ex]{\makecell[c]{\bf Pre-LLM\\ \bf Retention Ratio}}
& \multicolumn{2}{c}{\makecell{\bf Qwen2.5-Omni-3B/7B}}
& \multicolumn{2}{c}{\makecell{\bf MiniCPM-o-2.6}} \\
\cmidrule(lr){2-3}
\cmidrule(lr){4-5}
& \textit{Audio-intact} & \textit{Both-selected}
& \textit{Audio-intact} & \textit{Both-selected} \\
\midrule
25\% & 12\%-100\% & 20\%-55\% & 11\%-100\% & 20\%-50\% \\
20\% & 7\%-100\% & 15\%-50\% & 5\%-100\% & 15\%-45\% \\
15\% & -- & 10\%-45\% & -- & 10\%-40\% \\
10\% & -- & 6\%-35\% & -- & 5\%-35\% \\
\bottomrule
\end{tabular}}
\caption{Model-specific visual-audio allocations for each pre-LLM retention ratio $R$. Each entry gives $R_v$-$R_a$, the retained percentages of visual and audio tokens, whose token-count-weighted average approximately matches $R$. \textit{Audio-intact} keeps all audio tokens, whereas \textit{Both-selected} compresses both modalities.}
\label{tab:retention_ratio}
\end{table}
\paragraph{Retention Ratios and Configuration.} 
The modality-specific retention ratios differ between Qwen2.5-Omni-3B/7B and MiniCPM-o-2.6 because the two model families use different audio and video tokenization schemes. 
For each temporal window, Qwen2.5-Omni produces approximately 50 audio tokens and 288 visual tokens, whereas MiniCPM-o-2.6 produces approximately 750 audio tokens for every 4,096 visual tokens. 
We therefore allocate the audio and visual token budgets according to each model's native tokenization ratio, rather than applying identical modality-specific retention ratios across models, as detailed in Table~\ref{tab:retention_ratio}. 
This preserves a comparable balance between the two modalities under the same overall pre-LLM retention budget. 
For all models, the maximum pixel budget per sampled frame is set to 128$\times$28$\times$28. 

\subsection{DPC-KNN Formulation}
\label{app:dpc_knn}

For modality $m\in\{v,a\}$, let $\mathbf{X}^{m}=\{\mathbf{x}_{i}^{m}\}_{i=1}^{N_m}$ denote the input modality tokens and $K_m$ the pre-LLM token budget. 
After importance selection, the top $K_m^{\mathrm{imp}}$ tokens form $\mathcal{S}_{\mathrm{imp}}^m$. 
Coverage selection applies DPC-KNN~\cite{du2016study,rodriguez2014clustering} to the remaining candidates using the joint distance: 
\begin{equation}
d_{ij}^{m}
=
1-\cos\left(\mathbf{x}_i^m,\mathbf{x}_j^m\right)
+\lambda d_{ij}^{m,\mathrm{pos}}
\label{eq:appendix_dpc_distance}
\end{equation}
where $d_{ij}^{m,\mathrm{pos}}$ denotes the temporal distance for audio or spatiotemporal distance for video. 
We independently normalize the positional distances to $[0,1]$ within each modality. 

For each candidate token $i\notin\mathcal{S}_{\mathrm{imp}}^m$, let $\operatorname{KNN}_k^m(i)$ denote its $k$ nearest candidates excluding itself under $d_{ij}^m$. 
We set $k$=7 in all experiments.
Its local density is defined as: 
\begin{equation}
\rho_i^m
=
\exp\left(
-\frac{1}{k}
\sum_{j\in\operatorname{KNN}_k^m(i)}
\left(d_{ij}^m\right)^2
\right)
\label{eq:dpc_density}
\end{equation}

In the following, all indices $i$, $j$, and $q$ refer to candidates
outside $\mathcal{S}_{\mathrm{imp}}^m$. 
The separation from higher-density candidates is defined as: 
\begin{equation}
\delta_i^m
=
\begin{cases}
\displaystyle
\min_{j:\rho_j^m>\rho_i^m} d_{ij}^m,
&
\rho_i^m<\displaystyle\max_q\rho_q^m,
\\[6pt]
\displaystyle
\max_{j\neq i}d_{ij}^m,
&
\text{otherwise}
\end{cases}
\label{eq:dpc_separation}
\end{equation}

The representative score is computed as:
\begin{equation}
\gamma_i^m=\rho_i^m\delta_i^m
\label{eq:dpc_score}
\end{equation}
A high score indicates that a token is both representative of its local neighborhood and separated from other high-density regions. 
The top $K_m-K_m^{\mathrm{imp}}$ candidates form
$\mathcal{S}_{\mathrm{cov}}^m$. 
Combining them with the importance-selected tokens gives:
\begin{equation}
\mathcal{S}_{\mathrm{pre}}^m
=
\mathcal{S}_{\mathrm{imp}}^m
\cup
\mathcal{S}_{\mathrm{cov}}^m,
\quad
\left|\mathcal{S}_{\mathrm{pre}}^m\right|=K_m
\label{eq:appendix_pre_retained_set}
\end{equation}

\section{Reproduction Details}

For each target pre-LLM retention ratio $R$, we use the model-specific visual--audio allocations $R_v$--$R_a$ in Table~\ref{tab:retention_ratio}. 
Unless otherwise specified, the visual and audio retention ratios of all methods follow these allocations. 
FastV, VisionZip, FastVID, and VidCom$^2$ are evaluated under the \textit{Audio-intact} mode, while FastV-om, VisionZip-om, and OmniZip use the \textit{Both-selected} mode. 
For the visual-only methods, we keep $R_a=100\%$ and reallocate the remaining token budget to visual tokens for each example.

\noindent\textbf{FastV}~\cite{chen2024image} (ECCV 2024).
FastV prunes visual tokens after the $K$-th LLM block according to the attention scores from the last text token. 
We follow the official setting with $K$=2. 
FastV-om applies the same attention-based selection to both visual and audio tokens. 
At $R=$10\%, FastV-om is not reported for Qwen2.5-Omni-7B or MiniCPM-o-2.6 because their target visual retention ratios, 6\% and 5\%, respectively, are below the minimum layer-averaged retention of 2/28$\approx$7.1\% imposed by the first two unpruned blocks.

\noindent\textbf{VisionZip}~\cite{yang2025visionzip} (CVPR 2025).
VisionZip selects dominant visual tokens using encoder attention and optionally merges redundant tokens into contextual representatives. 
Since compression at the encoder output conflicts with visual pooling, we apply it after pooling following~\cite{shen2026fastvid}. 
We use two temporal grids per visual window and set the contextual-token ratio to $0$, allocating the entire budget to dominant tokens. 
VisionZip-om applies the same attention-based selection to audio tokens using two-second windows. 

\noindent\textbf{FastVID}~\cite{shen2026fastvid} (NeurIPS 2025).
FastVID dynamically segments video tokens according to adjacent-frame similarities and allocates the visual-token budget based on segment- and frame-level saliency. 
We apply it to pooled visual tokens and set the minimum number of segments to $c$=8, the segmentation threshold to $\tau$=0.84, and the anchor-frame step to $p$=4. 
Within each segment, the token budget of every non-anchor frame is further scaled by 0.7. 
Since this scaling is applied after budget allocation, the realized retention ratio can be slightly lower than the target $R$. 

\noindent\textbf{VidCom$^2$}~\cite{liu2025video} (EMNLP 2025).
VidCom$^2$ selects informative visual tokens by identifying feature outliers from video- and frame-level representations using multi-scale Gaussian similarities. 
We apply it to pooled visual tokens with a low-variance channel ratio of 0.5, a temperature of 0.01, and Gaussian scales $\{2^{-3},2^{-2},2^{-1},2^{0},2^{1}\}$.
Its per-frame retention ratios are dynamically adjusted to match the target visual-token budget. 

\noindent\textbf{OmniZip}~\cite{tao2026omnizip} (CVPR 2026).
OmniZip selects dominant audio tokens using audio-encoder attention and retains an additional 5\% contextual representatives, each aggregating at most $g$=3 discarded tokens.
The resulting audio retention guides the group-wise video discard ratios, which are normalized to the target visual budget and clipped to [35\%,75\%]. 
Within each video group, OmniZip applies DPC-KNN selection to alternating frames and inter-frame similarity-based selection to the remaining frames. 
We use the official implementation with a contextual ratio of 0.05 and $g$=3. 
Since the maximum video discard ratio is 75\%, $R_v$ cannot be reduced below 25\%. 
Therefore, at R=25\%, we use $R_v=R_a$=25\%. The audio budget consists of 20\% dominant tokens and 5\% contextual representatives, both of which are counted toward $R_a$.

\noindent\textbf{OmniSIFT}~\cite{ding2026omnisift} (ICML 2026).
OmniSIFT adopts modality-asymmetric token compression, using spatiotemporal pruning to remove intra- and inter-frame visual redundancy and vision-guided selection to filter audio tokens. 
The original method jointly optimizes these compression modules through alignment training. 
For a fair comparison with training-free methods, OmniSIFT\textsuperscript{$\circ$} uses only its token compression strategy without additional training. 

\noindent\textbf{SEATS}~\cite{xin2026stage}.
SEATS combines window-wise pre-LLM compression with progressive query-guided compression within the LLM. 
Before the LLM, it compresses visual and audio tokens within temporal windows. 
Within the LLM, it progressively removes tokens at predefined layers, followed by optional late-layer token removal. Since late-layer token removal further reduces FLOPs but may also degrade performance, we evaluate two variants for a fair comparison: SEATS\textsuperscript{$\dagger$} follows the original configuration, whereas SEATS\textsuperscript{$\star$} disables late-layer removal while retaining both pre-LLM and progressive inner-LLM compression. 
Both variants are configured to match the final modality-specific retention ratios of our method. 
SEATS adopts a depth-specific layer-selection schedule whose official configuration is provided only for 28-layer architectures, without an adaptation for the 36-layer Qwen2.5-Omni-3B. 
To avoid modifying this core strategy, we evaluate both variants only on Qwen2.5-Omni-7B and MiniCPM-o-2.6, which both contain 28 LLM layers. 

For a fair comparison, we reproduce all methods under a unified evaluation setting and evaluate them using NVIDIA H20 GPUs. 

\section{Computing Cost Evaluation}
Following prior work~\cite{tao2026omnizip}, we evaluate the computational efficiency of token compression using the floating-point operations (FLOPs) introduced by vision and audio tokens during the prefilling and decoding stages. 
Let $n=n^{(v)}+n^{(a)}$ denote the total number of vision and audio tokens, where $d$ is the hidden dimension and $m$ is the FFN intermediate dimension. Each transformer layer is approximated as a multi-head attention (MHA) module and a feed-forward network (FFN). Under the convention that one multiply--accumulate operation is counted as one FLOP, the prefilling cost per layer is approximated as $4nd^2+2n^2d+2ndm$. During autoregressive decoding, the computational cost of generating $\mathcal{R}$ tokens is approximated as $\mathcal{R}(4d^2+2dm)+2\sum_{i=1}^{\mathcal{R}}d(n+i)$. We set $\mathcal{R}$=100 in all experiments. Therefore, for an LLM with $T$ transformer layers and a constant sequence length $n$, the total FLOPs are formulated as: 
\begin{equation}
\begin{aligned}
\text{FLOPs}={}&T(4nd^2+2n^2d+2ndm)+{}\\
&T\mathcal{R}\left((4d^2+2dm)+2\left(dn+\frac{d(\mathcal{R}+1)}{2}\right)\right)
\end{aligned}
\label{eq:computing_cost}
\end{equation}

For methods whose sequence length changes across transformer layers, we apply Eq.~\eqref{eq:computing_cost} to each layer using its effective multimodal sequence length $n_i$ and sum the resulting costs over all layers. 
Pre-LLM compression methods use the compressed sequence throughout the transformer. 
FastV and FastV-om retain the full multimodal sequence in the first two layers and compress it in subsequent layers. SEATS$^{\star}$ progressively reduces tokens after layers 16, 18, and 20, whereas SEATS$^{\dagger}$ additionally discards the remaining multimodal tokens from layer 24 onward. 
Our method performs its second-stage compression at layer 18 for Qwen2.5-Omni-7B and MiniCPM-o-2.6, and at layer 26 for Qwen2.5-Omni-3B. 
FastVID additionally reduces non-key-frame tokens by a factor of 0.7. Vision and audio encoders, modality projectors, textual tokens, token-selection operations, and the language-model head are excluded. 

\begin{figure}[!t]
  \centering
  \includegraphics[width=1.0\linewidth]{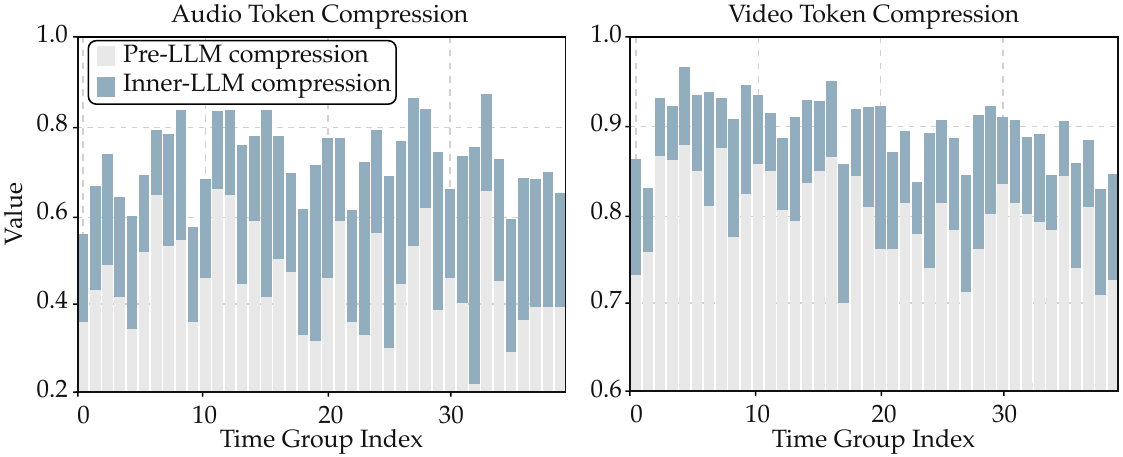}
  \caption{Visualization of the Pre-LLM and Inner-LLM token compression process.}
  \label{fig:compression_visualization}
\end{figure}

\section{More Experimental Details}

\paragraph{Different Backbones.} 
\begin{table*}[htbp]
\centering
\setlength{\tabcolsep}{3.2pt}
\renewcommand{\arraystretch}{0.92}
\small
\begin{tabular}{@{}lccc*{5}{c}cc@{}}
\toprule
\multirow{2}{*}{\textbf{Method}}
& \multicolumn{3}{c}{\textbf{Efficiency}}
& \multicolumn{5}{c}{\textbf{Benchmark Performance}}
& \multicolumn{2}{c}{\textbf{Average}} \\
\cmidrule(lr){2-4}
\cmidrule(lr){5-9}
\cmidrule(lr){10-11}
&
\makecell{\textbf{FLOPs (T)}}
&
\makecell{\textbf{Relative}\\\textbf{FLOPs}}
&
\makecell{\textbf{Retention}\\\textbf{Ratio}}
&
\textbf{AVUT}
&
\makecell{\textbf{World}\\\textbf{Sense}}
&
\makecell{\textbf{Daily}\\\textbf{Omni}}
&
\makecell{\textbf{Video}\\\textbf{MME}}
&
\makecell{\textbf{LVOmni}\\\textbf{Bench}}
&
\makecell{\textbf{Avg.}\\\textbf{Score}}
&
\makecell{\textbf{Relative}\\\textbf{Perf. (\%)}}
\\
\midrule

\rowcolor{gray!18}
Qwen2.5-Omni-3B
& 37.4 & 100.0\% & 100\%/100\%
& 62.5 & 46.5 & 61.0 & 62.5 & 35.2
& 53.5 & 100.0 \\

\midrule

Random
& 6.7 & 18.1\% & 25\%/--
& 56.3 & 42.7 & 54.5 & 61.1 & 33.6
& 49.6 & 92.7 \\

FastV-om
& 8.5 & 22.6\% & 100\%/25\%
& 55.4 & 43.3 & 54.2 & 60.3 & 32.9
& 49.2 & 92.0 \\

OmniZip
& 6.7 & 18.1\% & 25\%/--
& 54.4 & 42.4 & 51.8 & 61.1 & 32.8
& 48.5 & 90.7 \\

OmniSIFT\textsuperscript{$\circ$}
& 6.7 & 18.1\% & 25\%/--
& 57.0 & 42.6 & 50.5 & \textbf{61.8} & \textbf{34.3}
& \underline{49.2} & \underline{92.0} \\

\textbf{OmniPack (w/o M)}
& 6.7 & 18.1\% & 25\%/--
& \textbf{58.3} & \textbf{45.2} & \textbf{55.4} & \underline{61.7} & 33.6
& \textbf{50.8} & \textbf{95.0} \\

\rowcolor{blue!8}
\textbf{OmniPack}
& 5.8 & 15.5\% & 25\%/12.5\%
& \underline{57.9} & \underline{45.1} & \underline{55.3} & \underline{61.7} & \underline{33.8}
& \textbf{50.8} & \textbf{95.0} \\

\midrule

Random
& 5.3 & 14.2\% & 20\%/--
& \underline{55.7} & 42.4 & 51.3 & 60.1 & \textbf{34.6}
& 48.8 & 91.2 \\

FastV-om
& 7.1 & 18.9\% & 100\%/20\%
& 54.1 & \underline{42.7} & 51.6 & 57.6 & 31.1
& 47.4 & 88.6 \\

OmniSIFT\textsuperscript{$\circ$}
& 5.3 & 14.2\% & 20\%/--
& 54.2 & 39.9 & 51.0 & \textbf{61.1} & 33.3
& 47.9 & 89.5 \\

\textbf{OmniPack (w/o M)}
& 5.3 & 14.2\% & 20\%/--
& \textbf{58.2} & \textbf{44.4} & \textbf{54.2} & \textbf{61.1} & \underline{33.7} & \textbf{50.3} & \textbf{94.0} \\

\textbf{OmniPack}
& 4.6 & 12.2\% & 20\%/10\%
& \textbf{58.2} & \textbf{44.4} & \underline{53.9} & \underline{61.0} & 33.6
& \underline{50.2} & \underline{93.8} \\

\midrule

Random
& 2.6 & 7.0\% & 10\%/--
& 51.7 & 38.9 & 49.0 & 56.8 & 32.1
& 45.7 & 85.4 \\

FastV-om
& 4.5 & 12.2\% & 100\%/10\%
& 47.6 & 40.2 & 49.4 & 48.6 & 29.7
& 43.1 & 80.6 \\

VisionZip-om
& 2.6 & 7.0\% & 10\%/--
& 52.6 & 42.1 & \textbf{50.5} & 57.7 & 31.3
& 46.8 & 87.5 \\

OmniSIFT\textsuperscript{$\circ$}
& 2.6 & 7.0\% & 10\%/--
& 49.9 & 38.2 & 48.1 & 56.6 & 32.2
& 45.0 & 84.1 \\

\textbf{OmniPack (w/o M)}
& 2.6 & 7.0\% & 10\%/--
& \underline{55.6} & \textbf{42.8} & \underline{50.1}
& \textbf{58.9} & \textbf{32.8}
& \textbf{48.0} & \textbf{89.7} \\

\rowcolor{blue!8}
\textbf{OmniPack}
& 2.3 & 6.1\% & 10\%/5\%
& \textbf{55.7} & \underline{42.5} & 50.0
& \underline{58.6} & \underline{32.7}
& \underline{47.9} & \underline{89.5} \\

\midrule

\rowcolor{gray!18}
MiniCPM-o-2.6
& 73.2 & 100.0\% & 100\%/100\%
& 53.2 & 40.8 & 52.2 & 62.0 & 29.5
& 47.5 & 100.0 \\

\midrule

Random
& 14.9 & 20.4\% & 25\%/--
& 52.8 & 39.8 & 53.9 & 61.5 & \underline{35.0}
& 48.6 & 102.3 \\

FastV-om
& 19.1 & 26.1\% & 100\%/25\%
& 49.6 & 37.1 & 50.4 & 52.7 & 30.1
& 44.0 & 92.6 \\

VisionZip-om
& 14.9 & 20.4\% & 25\%/--
& 50.9 & 40.4 & 53.3 & 60.9 & \textbf{35.5}
& 48.2 & 101.5 \\

OmniSIFT\textsuperscript{$\circ$}
& 14.9 & 20.4\% & 25\%/--
& 50.6 & 40.4 & 52.3 & 60.0 & 33.2
& 47.3 & 99.6 \\

SEATS\textsuperscript{$\dagger$}
& 11.6 & 15.8\% & 25\%/12.5\%
& \underline{53.2} & 40.3 & 53.4 & 60.6 & 34.9
& 48.5 & 102.1 \\

SEATS\textsuperscript{$\star$}
& 12.5 & 17.1\% & 25\%/12.5\%
& 52.9 & 40.5 & 53.1 & 62.1 & 33.9
& 48.5 & 102.1 \\

\textbf{OmniPack (w/o M)}
& 14.9 & 20.4\% & 25\%/--
& 52.9 & \underline{40.8} & \underline{54.0} & \textbf{63.5} & 34.8
& \underline{49.2} & \underline{103.6} \\

\rowcolor{blue!8}
\textbf{OmniPack}
& 12.2 & 16.7\% & 25\%/12.5\%
& \textbf{53.8} & \textbf{40.9} & \textbf{54.1} & \underline{63.4} & 34.2 & \textbf{49.3} & \textbf{103.8} \\

\midrule

Random
& 11.8 & 16.2\% & 20\%/--
& 51.2 & 39.2 & 53.1 & 61.3 & 35.0
& 48.0 & 101.1 \\

FastV-om
& 16.2 & 22.2\% & 100\%/20\%
& 48.0 & 36.6 & 48.4 & 53.6 & 29.9
& 43.3 & 91.2 \\

VisionZip-om
& 11.8 & 16.2\% & 20\%/--
& 49.9 & 39.7 & 53.0 & 61.1 & 35.2
& 47.8 & 100.6 \\

OmniSIFT\textsuperscript{$\circ$}
& 11.8 & 16.2\% & 20\%/--
& 50.5 & 39.0 & 51.6 & 60.0 & 33.3
& 46.9 & 98.7 \\

SEATS\textsuperscript{$\dagger$}
& 9.2 & 12.6\% & 20\%/10\%
& 50.6 & 40.1 & 53.0 & 60.8 & \textbf{36.3}
& 48.2 & 101.5 \\

SEATS\textsuperscript{$\star$}
& 10.0 & 13.6\% & 20\%/10\%
& 50.7 & \underline{40.3} & 52.5 & 61.7 & 32.9
& 47.6 & 100.2 \\

\textbf{OmniPack (w/o M)}
& 11.8 & 16.2\% & 20\%/--
& \underline{52.6} & \textbf{40.5} & \textbf{53.6} & \textbf{63.7} & 35.4
& \textbf{49.2} & \textbf{103.6} \\

\rowcolor{blue!8}
\textbf{OmniPack}
& 9.8 & 13.3\% & 20\%/10\%
& \textbf{52.8} & \underline{40.3} & \underline{53.5} & \underline{63.0} & \underline{35.8}
& \underline{49.1} & \underline{103.4} \\

\midrule

Random
& 6.0 & 8.2\% & 10\%/--
& 46.0 & 37.1 & 47.0 & 58.8 & 32.8
& 44.3 & 93.3 \\

VisionZip-om
& 6.0 & 8.2\% & 10\%/--
& 45.5 & \underline{37.6} & 47.3 & 58.1 & \underline{33.8}
& 44.5 & 93.7 \\

OmniSIFT\textsuperscript{$\circ$}
& 6.0 & 8.2\% & 10\%/--
& 45.1 & 36.5 & 47.6 & 57.1 & \textbf{34.2}
& 44.1 & 92.8 \\

SEATS\textsuperscript{$\dagger$}
& 4.7 & 6.5\% & 10\%/5\%
& 46.9 & \underline{37.6} & 47.8 & 57.8 & 33.6
& 44.7 & 94.1 \\

SEATS\textsuperscript{$\star$}
& 5.1 & 7.0\% & 10\%/5\%
& \underline{47.1} & 37.5 & 46.3 & 58.1 & 33.6
& 44.5 & 93.7 \\

\textbf{OmniPack (w/o M)}
& 6.0 & 8.2\% & 10\%/--
& \textbf{50.1} & \textbf{38.7} & \underline{48.6}
& \underline{60.1} & \underline{33.8}
& \textbf{46.3} & \textbf{97.5} \\

\rowcolor{blue!8}
\textbf{OmniPack}
& 5.0 & 6.8\% & 10\%/5\%
& 47.0 & \textbf{38.7} & \textbf{48.7}
& \textbf{61.2} & 33.6
& \underline{45.8} & \underline{96.4} \\

\bottomrule
\end{tabular}
\caption{Cross-backbone comparison on Qwen2.5-Omni-3B and MiniCPM-o-2.6.}
\label{tab:main_results_3b_minicpm_25_20_10}
\end{table*}
Beyond the 15\% retention setting reported in the main paper, Table~\ref{tab:main_results_3b_minicpm_25_20_10} provides additional results at 25\%, 20\%, and 10\% pre-LLM retention ratios on Qwen2.5-Omni-3B and MiniCPM-o-2.6. 
On Qwen2.5-Omni-3B, OmniPack retains 95.0\%, 93.8\%, and 89.5\% of the original performance while using only 15.5\%, 12.2\%, and 6.1\% of the original FLOPs, respectively. 
On MiniCPM-o-2.6, OmniPack achieves 103.8\% and 103.4\% of the original performance at 25\% and 20\% retention, and retains 96.4\% at 10\% retention with only 6.8\% of the original FLOPs.
These results demonstrate a consistent performance--efficiency trade-off across different model scales, architectures, and compression budgets.

\begin{table*}[htbp]
\centering
\setlength{\tabcolsep}{3.2pt}
\renewcommand{\arraystretch}{0.92}
\small
\resizebox{1.0\linewidth}{!}{
\begin{tabular}{@{}lccc*{6}{c}c@{}}
\toprule
\multirow{2}{*}{\textbf{Method}}
& \multicolumn{3}{c}{\textbf{Efficiency}}
& \multicolumn{6}{c}{\textbf{Fine-Grained Performance}}
& \multirow{2}{*}{\textbf{Avg.}} \\
\cmidrule(lr){2-4}
\cmidrule(lr){5-10}
&
\makecell{\textbf{FLOPs (T)}}
&
\makecell{\textbf{Relative}\\\textbf{FLOPs}}
&
\makecell{\textbf{Retention}\\\textbf{Ratio}}
&
\makecell{\textbf{Audio Event}\\\textbf{Location}}
&
\makecell{\textbf{Audio Object}\\\textbf{Matching}}
&
\makecell{\textbf{Audio OCR}\\\textbf{Matching}}
&
\makecell{\textbf{Audio Information}\\\textbf{Extraction}}
&
\makecell{\textbf{Audio Character}\\\textbf{Matching}}
&
\makecell{\textbf{Audio Content}\\\textbf{Counting}}
&
\\
\midrule
\rowcolor{gray!18}
Qwen2.5-Omni-7B & 73.2 & 100.0\% & 100\% / 100\% & 35.9 & 67.2 & 58.2 & 85.9 & 68.1 & 40.7 & 64.0 \\
\midrule
Random & 14.9 & 20.4\% & 25\% / -- & 32.9 & 53.3 & 61.4 & 82.8 & 59.5 & \textbf{36.4} & 58.7 \\
FastV-om & 19.1 & 26.1\% & 100\% / 25\% & \underline{38.2} & 52.3 & \textbf{68.8} & 83.4 & 55.2 & \underline{35.6} & 59.3 \\
VisionZip-om & 14.9 & 20.4\% & 25\% / -- & 36.5 & 52.0 & 63.7 & 82.2 & 59.2 & 33.1 & 58.7 \\
OmniZip & 14.9 & 20.4\% & 25\% / -- & 37.0 & 53.8 & 64.3 & 70.8 & 59.7 & 27.9 & 56.9 \\
OmniSIFT\textsuperscript{$\circ$} & 14.9 & 20.4\% & 25\% / -- & 37.6 & 51.0 & 67.2 & 80.4 & 56.8 & 33.9 & 58.4 \\
SEATS\textsuperscript{$\dagger$} & 11.6 & 15.8\% & 25\% / 12.5\% & 36.5 & \underline{54.6} & 66.6 & \textbf{84.0} & 58.0 & \underline{35.6} & 60.0 \\
SEATS\textsuperscript{$\star$} & 12.5 & 17.1\% & 25\% / 12.5\% & 35.9 & \textbf{54.9} & 66.6 & \textbf{84.0} & 58.3 & \underline{35.6} & 60.1 \\
\addlinespace[1pt]
\textbf{OmniPack (w/o M)} & 14.9 & 20.4\% & 25\% / -- & \textbf{39.4} & 53.3 & 66.2 & \underline{83.4} & \underline{61.4} & \textbf{36.4} & \underline{60.7} \\
\rowcolor{blue!8} \textbf{OmniPack} & 12.2 & 16.7\% & 25\% / 12.5\% & \textbf{39.4} & 53.6 & \underline{67.5} & \underline{83.4} & \textbf{61.6} & \underline{35.6} & \textbf{61.0} \\
\bottomrule
\end{tabular}}
\caption{Fine-grained performance comparison on six AVUT subtasks using Qwen2.5-Omni-7B.}
\label{tab:avut_fine_grained_25}
\end{table*}

\begin{table*}[!t]
\centering
\setlength{\tabcolsep}{3.2pt}
\renewcommand{\arraystretch}{0.92}
\small
\resizebox{\linewidth}{!}{
\begin{tabular}{@{}lccc*{8}{c}c@{}}
\toprule
\multirow{2}{*}{\textbf{Method}}
& \multicolumn{3}{c}{\textbf{Efficiency}}
& \multicolumn{8}{c}{\textbf{Fine-Grained Performance}}
& \multirow{2}{*}{\textbf{Avg.}} \\
\cmidrule(lr){2-4}
\cmidrule(lr){5-12}
&
\makecell{\textbf{Prefill}\\\textbf{FLOPs (T)}}
&
\makecell{\textbf{Relative}\\\textbf{FLOPs}}
&
\makecell{\textbf{Retention}\\\textbf{Ratio}}
&
\makecell{\textbf{Tech \&}\\\textbf{Science}}
&
\makecell{\textbf{Culture \&}\\\textbf{Politics}}
&
\makecell{\textbf{Daily}\\\textbf{Life}}
&
\makecell{\textbf{Film \&}\\\textbf{TV}}
&
\textbf{Performance}
&
\textbf{Games}
&
\textbf{Sports}
&
\textbf{Music}
&
\\
\midrule

\rowcolor{gray!18}
Qwen2.5-Omni-7B
& 73.2 & 100.0\% & 100\%  / 100\%
& 52.4 & 51.1 & 48.3 & 44.1 & 42.0 & 41.0 & 41.7 & 47.3
& 46.6 \\

\midrule

Random
& 14.9 & 20.4\% & 25\% / --
& 48.8 & 44.7 & 44.5 & \underline{41.7} & 38.2 & 36.9 & 39.1 & 45.1
& 43.1 \\

FastV
& 19.1 & 26.1\% & 100\%/25\%
& 47.8 & 44.7 & 42.4 & 39.6 & 39.3 & 36.5 & 37.9 & 44.1
& 42.0 \\

FastV-om
& 19.1 & 26.1\% & 100\%/25\%
& \underline{49.6} & 46.3 & 43.8 & 40.9 & 39.0 & 37.3 & 39.8 & 44.8
& 43.3 \\

VisionZip
& 14.9 & 20.4\% & 25\% / --
& \underline{49.6} & 48.2 & 44.2 & 40.4 & \underline{42.7} & 40.8 & 38.8 & 45.6
& 44.0 \\

VisionZip-om
& 14.9 & 20.4\% & 25\% / --
& 48.4 & \underline{50.5} & 46.2 & 39.8 & 40.1 & \textbf{41.6} & 40.5 & \underline{47.5}
& 44.7 \\

FastVID
& 13.6 & 18.5\% & 25\% / --
& 47.4 & 45.3 & 43.0 & 40.6 & 40.8 & 37.8 & 39.3 & 45.1
& 42.8 \\

VidCom$^2$
& 14.9 & 20.4\% & 25\% / --
& 48.2 & 47.3 & 45.0 & 39.8 & 39.3 & 37.8 & 40.0 & 44.1
& 43.3 \\

OmniZip
& 14.9 & 20.4\% & 25\% / --
& 47.8 & 44.0 & 43.6 & 38.5 & 37.5 & \underline{41.2} & 38.4 & 46.3
& 42.6 \\

OmniSIFT\textsuperscript{$\circ$}
& 14.9 & 20.4\% & 25\% / --
& 45.9 & 46.6 & 44.5 & 37.5 & 37.8 & 36.9 & 38.6 & 43.6
& 42.1 \\

SEATS\textsuperscript{$\dagger$}
& 11.6 & 15.8\% & 25\%/12.5\%
& 48.2 & \textbf{51.1} & \textbf{46.8} & 40.6
& 40.8 & 39.5 & \textbf{41.6} & 46.8
& \underline{45.0} \\

SEATS\textsuperscript{$\star$}
& 12.5 & 17.1\% & 25\%/12.5\%
& 48.4 & \textbf{51.1} & \textbf{46.8} & 41.4
& 40.8 & 39.1 & \underline{41.2} & 46.8
& \underline{45.0} \\

\textbf{OmniPack (w/o M)}
& 14.9 & 20.4\% & 25\% / --
& \textbf{49.8} & 49.8 & \underline{46.5} & \textbf{43.0}
& \textbf{43.0} & 38.2 & 39.8 & \textbf{48.3}
& \textbf{45.3} \\

\rowcolor{blue!8}
\textbf{OmniPack}
& 12.2 & 16.7\% & 25\%/12.5\%
& 49.4 & 50.2 & 45.4 & \textbf{43.0}
& \underline{42.7} & 39.5 & 40.0 & \underline{47.5}
& \textbf{45.3} \\

\bottomrule
\end{tabular}
}
\caption{Fine-grained performance comparison on eight WorldSense categories using Qwen2.5-Omni-7B.} 
\label{tab:worldsense_fine_grained_25}
\end{table*}
\begin{table*}[!t]
\centering
\setlength{\tabcolsep}{3.2pt}
\renewcommand{\arraystretch}{0.92}
\small
\resizebox{1.0\linewidth}{!}{
\begin{tabular}{@{}lccc*{6}{c}c@{}}
\toprule
\multirow{2}{*}{\textbf{Method}}
& \multicolumn{3}{c}{\textbf{Efficiency}}
& \multicolumn{6}{c}{\textbf{Fine-Grained Performance}}
& \multirow{2}{*}{\textbf{Avg.}} \\
\cmidrule(lr){2-4}
\cmidrule(lr){5-10}
&
\makecell{\textbf{FLOPs (T)}}
&
\makecell{\textbf{Relative}\\\textbf{FLOPs}}
&
\makecell{\textbf{Retention}\\\textbf{Ratio}}
&
\makecell{\textbf{Event}\\\textbf{Sequence}}
&
\makecell{\textbf{AV Event}\\\textbf{Alignment}}
&
\textbf{Inference}
&
\textbf{Reasoning}
&
\makecell{\textbf{Context}\\\textbf{Understanding}}
&
\textbf{Comparative}
&
\\
\midrule
\rowcolor{gray!18}
Qwen2.5-Omni-7B
& 73.2 & 100.0\% & 100\% / 100\%
& 58.2 & 52.5 & 76.6 & 78.9 & 59.1 & 70.2
& 63.9 \\

\midrule

Random
& 14.9 & 20.4\% & 25\% / --
& 49.4 & 41.6 & 73.4 & 73.7 & \underline{52.3} & \underline{69.5}
& 57.1 \\

FastV-om
& 19.1 & 26.1\% & 100\%/25\%
& 50.3 & 44.5 & 78.6 & \textbf{77.1} & \underline{52.3} & 64.9
& 58.7 \\

VisionZip-om
& 14.9 & 20.4\% & 25\% / --
& 50.3 & 45.8 & 77.3 & 74.3 & 49.7 & 67.2
& 58.2 \\

OmniZip
& 14.9 & 20.4\% & 25\% / --
& 50.7 & 41.6 & 71.4 & 72.0 & 47.7 & 66.4
& 55.9 \\

OmniSIFT\textsuperscript{$\circ$}
& 14.9 & 20.4\% & 25\% / --
& 49.7 & 43.3 & 77.3 & 72.6 & \underline{52.3} & \textbf{70.2}
& 58.0 \\

SEATS\textsuperscript{$\dagger$}
& 11.6 & 15.8\% & 25\%/12.5\%
& \textbf{53.9} & 46.2 & 76.6 & 75.4 & 51.8 & 67.9
& 59.7 \\

SEATS\textsuperscript{$\star$}
& 12.5 & 17.1\% & 25\%/12.5\%
& 52.9 & 47.1 & 77.9 & \underline{76.6} & 51.3 & 67.9
& 59.8 \\

\textbf{OmniPack (w/o M)}
& 14.9 & 20.4\% & 25\% / --
& \textbf{53.9} & \underline{47.5} & \underline{79.8}
& 74.3 & \underline{52.3} & \underline{68.7}
& \underline{60.3} \\

\rowcolor{blue!8}
\textbf{OmniPack}
& 12.2 & 16.7\% & 25\%/12.5\%
& \underline{53.6} & \textbf{48.7} & \textbf{80.5}
& 74.3 & \textbf{52.9} & 67.9
& \textbf{60.6} \\
\bottomrule
\end{tabular}}
\caption{Fine-grained performance comparison on six DailyOmni subtasks using Qwen2.5-Omni-7B.}
\label{tab:dailyomni_fine_grained_25}
\end{table*}
\begin{table}[htbp]
\centering
\setlength{\tabcolsep}{4.2pt}
\renewcommand{\arraystretch}{0.92}
\small
\resizebox{\linewidth}{!}{
\begin{tabular}{@{}lcc*{3}{c}c@{}}
\toprule
\multirow{2}{*}{\textbf{Method}}
& \multicolumn{2}{c}{\textbf{Efficiency}}
& \multicolumn{3}{c}{\textbf{Fine-Grained Performance}}
& \multirow{2}{*}{\textbf{Avg.}} \\
\cmidrule(lr){2-3}
\cmidrule(lr){4-6}
&
\makecell{\textbf{FLOPs (T)}}
&
\makecell{\textbf{Retention}\\\textbf{Ratio}}
&
\textbf{Short}
&
\textbf{Medium}
&
\textbf{Long}
&
\\
\midrule
\rowcolor{gray!18}
Qwen2.5-Omni-7B & 73.2 & 100\% / 100\% & 77.1 & 63.4 & 52.8 & 64.4 \\
\midrule
Random & 8.9 & 15\% / -- & 69.8 & 63.8 & 55.4 & 63.0 \\
FastV-om & 13.5 & 100\% / 15\% & 62.2 & 59.1 & 51.2 & 57.5 \\
VisionZip-om & 8.9 & 15\% / -- & 68.3 & 62.9 & 53.6 & 61.6 \\
OmniSIFT\textsuperscript{$\circ$} & 8.9 & 15\% / -- & 69.7 & \underline{65.1} & 53.8 & 62.9 \\
SEATS\textsuperscript{$\dagger$} & 6.9 & 15\% / 7.5\% & 71.6 & \underline{65.1} & 54.2 & 63.6 \\
SEATS\textsuperscript{$\star$} & 7.5 & 15\% / 7.5\% & 71.8 & \underline{65.1} & 54.4 & 63.8 \\
\addlinespace[1pt]
\textbf{OmniPack (w/o M)} & 8.9 & 15\% / -- & \underline{72.4} & \textbf{65.2} & \underline{56.4} & \textbf{64.7} \\
\rowcolor{blue!8}
\textbf{OmniPack} & 7.3 & 15\% / 7.5\% & \textbf{72.6} & \underline{65.1} & \textbf{56.7} & \textbf{64.7} \\
\bottomrule
\end{tabular}}
\caption{Duration-wise performance comparison on VideoMME using Qwen2.5-Omni-7B .}
\label{tab:videomme_duration_15}
\end{table}

\paragraph{Fine-Grained Results.}
Tables~\ref{tab:avut_fine_grained_25}, \ref{tab:worldsense_fine_grained_25}, \ref{tab:dailyomni_fine_grained_25}, and \ref{tab:videomme_duration_15} provide fine-grained comparisons across AVUT, WorldSense, DailyOmni, and VideoMME. 
At the 25\%/12.5\% retention setting, OmniPack performs particularly well in audio-event localization and character matching on AVUT, as well as audio-visual event alignment, inference, and context understanding on DailyOmni. 
These tasks require associating acoustic evidence with relevant visual content, demonstrating that our audio-visual collaboration preserves complementary cross-modal cues under aggressive compression. 
The consistent performance across diverse WorldSense categories further shows that this benefit generalizes across different video domains. 
On VideoMME, OmniPack performs best on both short- and long-duration videos and remains competitive on medium-duration videos at the 15\%/7.5\% retention setting. 
In particular, it retains 107.4\% of the original long-video performance, indicating that removing redundant audio-visual tokens is especially beneficial for long sequences. 

\paragraph{Compression Strategies Collaboration.} 
As illustrated in Figure~\ref{fig:compression_visualization}, pre-LLM compression removes modality-specific redundancy, while inner-LLM compression refines tokens with task-conditioned representations. 
Their contributions vary across temporal groups, with audio varying more due to sparse acoustic events and video remaining stable because of continuous visual content. 
This complementary behavior enables early compression followed by semantic refinement after audio-visual interaction. 

\begin{figure}[!t]
  \centering
  \includegraphics[width=1.00\linewidth]{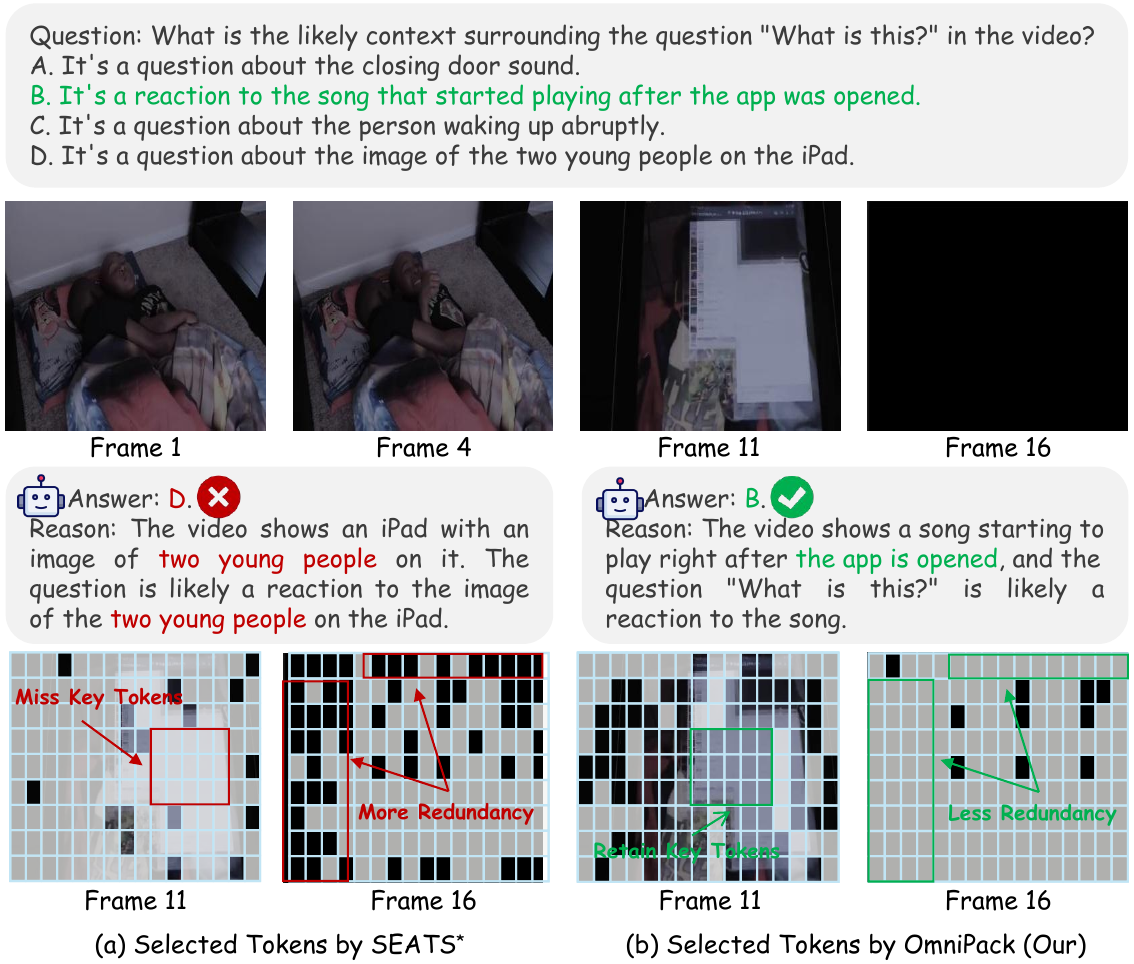}
  \caption{Qualitative comparison at the 25\% pre-LLM retention ratio. (a) SEATS\textsuperscript{$\star$} misses key tokens and answers incorrectly. (b) OmniPack preserves them and answers correctly.}
  \label{fig:case_studty}
\end{figure}

\paragraph{Case Study.} 
Figure~\ref{fig:case_studty} compares SEATS\textsuperscript{$\star$} and OmniPack under the same 25\% pre-LLM retention ratio. 
SEATS may discard critical evidence while retaining redundant tokens, resulting in an incorrect prediction. 
In contrast, OmniPack combines structure-aware selection with delayed query-conditioned audio-visual compression to better preserve task-relevant evidence and remove redundancy. 
This produces a more compact and informative representation, enabling the correct prediction under aggressive compression. 

\begin{figure}[!t]
  \centering
  \includegraphics[width=0.78\linewidth]{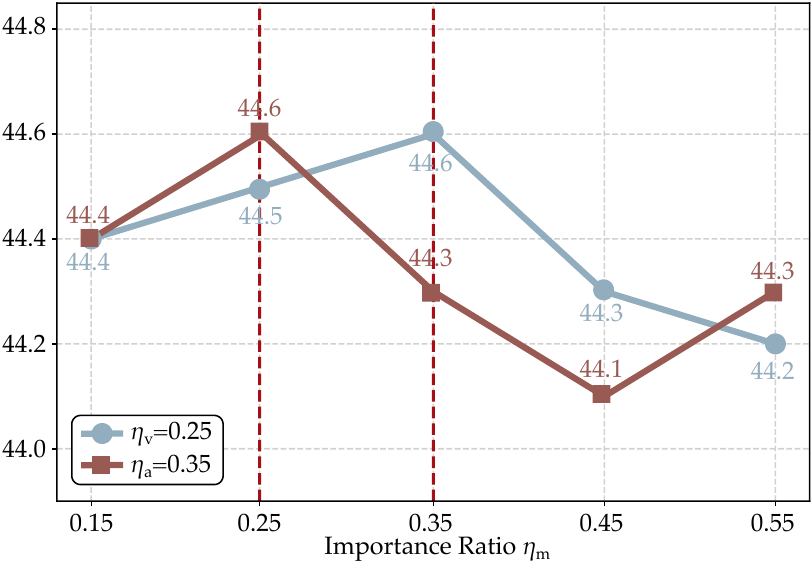}
  \caption{Sensitivity analysis of the importance ratios $\eta_m$.}
  \label{fig:importance_ratio}
\end{figure}

\paragraph{Parameter Ablation.}
As shown in Figure~\ref{fig:importance_ratio} and Table~\ref{tab:parameter_ablation}, we analyze the sensitivity of the importance ratios $(\eta_v,\eta_a)$, positional-distance weight $\lambda$, merging thresholds $(\tau_v,\tau_a)$, and merging weight $\zeta$. 
The importance-ratio analysis shows that $(\eta_v,\eta_a)$=(0.25,0.35) provides a favorable balance between the two modalities. 
Across the three benchmarks, varying $\lambda$ results in at most a 1.0-point difference, while the variations caused by $(\tau_v,\tau_a)$ and $\zeta$ remain within 0.5 and 0.2 points, respectively. 
Moreover, all tested configurations consistently outperform existing methods, demonstrating that our gains do not rely on carefully tuned hyperparameters. 
We therefore adopt $\lambda$=0.20, $(\tau_v,\tau_a)$=(0.10,0.05), and $\zeta$=0.10 as a balanced configuration across different benchmarks.
\begin{table}[t]
\centering
\begin{tabular}{@{}cc|ccc@{}}
\toprule
\textbf{Parameter}
&
\textbf{Value}
&
\textbf{AVUT}
&
\makecell{\textbf{World}\\\textbf{Sense}}
&
\makecell{\textbf{Daily}\\\textbf{Omni}}
\\
\midrule

\multirow{5}{*}{$\lambda$}
& 0.10 & 58.0 & 45.1 & 58.2 \\
& \textbf{0.20} & 58.1 & 44.6 & 58.2 \\
& 0.30 & 58.0 & 44.8 & 58.2 \\
& 0.40 & 58.1 & 44.5 & 57.7 \\
& 0.50 & 58.5 & 44.1 & 57.6 \\

\midrule

\multirow{5}{*}{$(\tau_v,\tau_a)$}
& (0.05,0.025) & 58.2 & 44.6 & 57.7 \\
& \textbf{(0.10,0.05)} & 58.1 & 44.6 & 58.2 \\
& (0.15,0.075) & 57.8 & 44.5 & 58.1 \\
& (0.20,0.10) & 58.0 & 44.7 & 58.0 \\
& (0.25,0.125) & 57.8 & 44.8 & 58.1 \\

\midrule

\multirow{5}{*}{$\zeta$}
& 0.05 & 58.1 & 44.5 & 58.3 \\
& \textbf{0.10} & 58.1 & 44.6 & 58.2 \\
& 0.15 & 57.9 & 44.7 & 58.1 \\
& 0.20 & 57.9 & 44.5 & 58.1 \\
& 0.25 & 57.9 & 44.6 & 58.1 \\

\bottomrule
\end{tabular}
\caption{Ablation of compression parameters. Each parameter is varied independently while the others remain fixed at their default values. Bold values indicate the default configuration.}
\label{tab:parameter_ablation}
\end{table}

\end{document}